%% file: main.tex
\documentclass[letterpaper, 10 pt, journal,twoside]{IEEEtran}
\IEEEoverridecommandlockouts
\input{configs}

\begin{document}
\title{Object Gathering with a Tethered Robot Duo}
\author{Yao Su$^{1}$\quad{}Yuhong Jiang$^{1}$\quad{}Yixin Zhu$^{1,2}$\quad{}Hangxin Liu$^{1}$
\thanks{Manuscript received: September 9, 2021; Revised: November 25, 2021; Accepted: December 22, 2021. This paper was recommended for publication by Editor Lucia Pallottino upon evaluation of the Associate Editor and Reviewers' comments. (\textit{Corresponding author: Hangxin Liu})}
\thanks{$^{1}$ Beijing Institute for General Artificial Intelligence (BIGAI).
}%
\thanks{$^{2}$ Institute for Artificial Intelligence, Peking University.
}%
\thanks{Emails: \tt{\{suyao, jiangyuhong, y, liuhx\}@bigai.ai}}%
\thanks{Digital Object Identifier (DOI): see top of this page.}}

\markboth{IEEE Robotics and Automation Letters. Preprint Version. Accepted December, 2021}
{Su \MakeLowercase{\textit{et al.}}: Object Gathering with a Tethered Robot Duo} 

\maketitle

\begin{abstract}
We devise a cooperative planning framework to generate optimal trajectories for a robot duo tethered by a flexible net to gather scattered objects spread in a large area. Specifically, the proposed planning framework first produces a set of dense waypoints for each robot, serving as the initialization for optimization. Next, we formulate an iterative optimization scheme to generate smooth and collision-free trajectories while ensuring cooperation within the robot duo to gather objects efficiently and avoid obstacles properly. We validate the generated trajectories in simulation and implement them in physical robots using \ac{mrac} to handle unknown dynamics of carried payloads. In a series of studies, we find that: (i) a U-shape cost function for maintaining separation distance is effective in planning cooperative robot duo, and (ii) the task efficiency is not always proportional to the tethered net's length. Given an environment configuration, our framework can gauge the optimal net length. To our best knowledge, ours is the first that provides such estimation for tethered robot duo.
\end{abstract}

\section{Introduction}

\begin{figure}[t!]
    \centering    
    \begin{subfigure}[b]{0.25\linewidth}
        \centering
        \includegraphics[width=\linewidth]{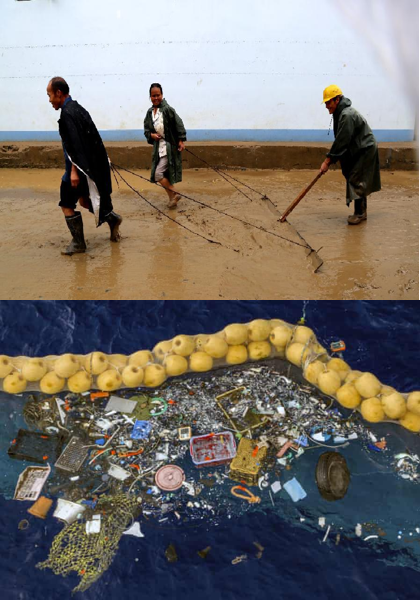}
        \caption{}
        \label{fig:moti_task}
    \end{subfigure}%
    \begin{subfigure}[b]{0.74\linewidth}
        \centering
        \includegraphics[width=\linewidth]{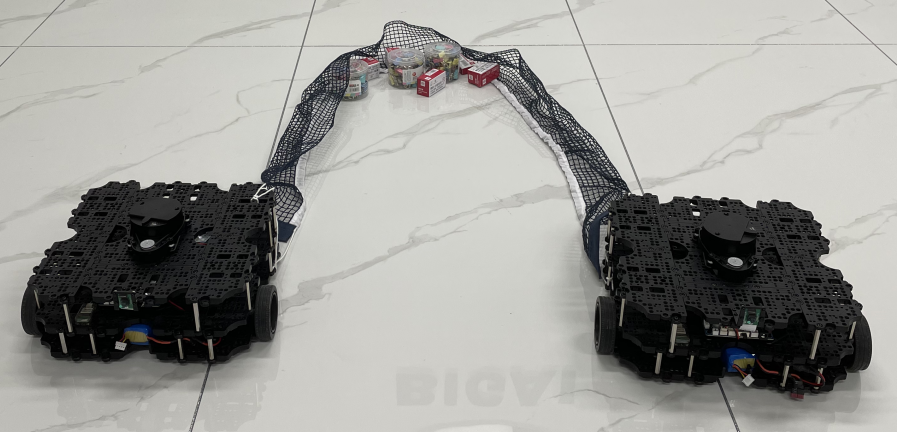}
        \caption{}
        \label{fig:moti_setup}
    \end{subfigure}%
    \caption{(a) \textbf{Two examples in daily life.} Humans use long/flexible structures to gather scattered objects, a similar setup akin to the one proposed in our work (Source: online). (b) \textbf{Two \acfp{wmr} tethered by a net as a duo.} Our framework \textbf{cooperatively} plans the robot duo's trajectories, indirectly maintaining the net's shape through motions. It also handles the \textbf{uncertainty} of the increasing payloads using \acf{mrac} to properly track the trajectories during operation.}
    \label{fig:motivation}
\end{figure}

\IEEEPARstart{W}e consider the task of gathering and carrying objects scattered on the floor by deploying two \acp{wmr} tethered by a flexible net or rope as a duo. Compared to picking and placing individual objects one by one, this setting is more intriguing and compelling for robots to collect small items spread across a large area autonomously. \cref{fig:moti_task} showcases two similar tasks in the physical world that share a similar spirit, wherein humans adopt a net to collect scattered objects with high efficiency.

The tethered robot duo must resolve two challenges. First, how to effectively generate \textbf{cooperative} trajectories spanned between two individuals? Despite a robot duo could accomplish more complex and dexterous tasks than a single robot~\cite{gulzar2018multi}, the physical connection among it also introduces additional constraints in planning. Compared with prior work concerning rigid connections (\eg, \cite{yamashita2003motion,heshmati2021predictive}), planning for the robot duo with a non-rigid connection (\eg, a rope or a net) is more challenging~\cite{bhattacharya2015topological,teshnizi2021motion}; one has to consider the task goals, obstacle avoidance, and net shape maintenance through individual's behaviors. Second, the tethered robot duo is subjected to \textbf{increasing payloads} during operation. In addition to the friction force introduced by the net, the dragging force increases significantly and deteriorates the trajectory tracking of each robot as the task progresses and more objects are gathered. Of note, this challenge is mostly overlooked in literature~\cite{bhattacharya2015topological,teshnizi2021motion,d2021catenary,cardona2021non}.

In this paper, we devise an optimization-based cooperation framework for a pair of tethered \acp{wmr} (\ie, a duo) to gather and transport objects with a net; see \cref{fig:moti_setup} for the setup. The core of our framework is a trajectory optimization scheme that produces smooth trajectories for the robot duo, which jointly accounts for task goals, obstacle avoidance, and net shape maintenance. A set of waypoints extended from a centerline connecting all the target objects along the robots' initial and final configurations serve as the initialization for the optimization scheme. An \ac{mrac} controller is further implemented to track the produced trajectories under unknown increasing payloads stably. Our framework is validated in both simulation and experiment. 

\subsection{Related Work}\label{sec:related}

Although \textbf{cooperative planning} has incubated many successful applications, such as search and rescue~\cite{furukawa2006recursive}, exploration~\cite{hussein2007effective}, object manipulation~\cite{yan2021decentralized,ritz2012cooperative}, payload transportation~\cite{zhang2021self,fink2011planning}, the planning problem of tethered robot duo to date is less explored. Prior work~\cite{kim2013topological,bhattacharya2015topological} mostly formulates it as a separation problem: The goal is to find trajectories for the robot duo to separate all target objects from obstacles. In such a scheme, prior methods assume an infinite separation distance between the robots, unpractical and unrealistic in many cases. Although Teshnizi \etal~\cite{teshnizi2021motion} relax this assumption and introduce a modified A$^\star$ planning algorithm given specific tether lengths, it fails to properly account for the obstacle avoidance (\eg, in final configuration) and net shape maintenance. We fully address these challenges in a new framework.

Gathering objects using the robot duo necessitates maintaining the shape of the flexible tether connecting the robots, though not necessarily precisely. Several setups have showcased the ability to \textbf{handle flexible connections} in diverse scenarios among robots. For instance, a dual-arm robot manipulates rope-like objects in clutter~\cite{mitrano2021learning}, three quadcopters connected by a net catch and throw a ball~\cite{ritz2012cooperative}, multiple robots manipulate deformable bed sheet~\cite{alonso2015local}, a visual servoing scheme maintains the tether's shape between two-wheel robots~\cite{laranjeira2017catenary,laranjeira2020catenary}, and a pair of tethered quadcopters move an object using the tether~\cite{d2021catenary,cardona2021non}. Although the above literature strongly suggests the significance of maintaining tether/net shapes in cooperative planning, directly and precisely controlling its shape is oftentimes unnecessary, resulting in increased complexity in both modeling and control. To tackle this problem, our proposed framework adopts a U-shape cost function to maintain a proper distance between the two tethered robots, such that the net shape is indirectly controlled.

As an essential branch of robust control, \textbf{\acf{mrac}} can maintain the platform's stability with unmodeled dynamics. It has been implemented on various robot platforms with inaccurate physical parameters, unknown payloads, or external disturbance~\cite{lavretsky2013robust}. For example, a tilt-rotor quadcopter platform using \ac{mrac} pulls an unmodeled cart~\cite{anderson2020constrained}, and \acp{wmr} adopts \ac{mrac} maintain the platform stability with inaccurate physical parameters~\cite{canigur2012model}. Similarly, we implement \ac{mrac} on the tethered robot duo for the object gathering task, since the physical parameters of the objects are unknown, and the drag force increases with more objects collected by the net.

\subsection{Overview}

We organize the remainder of the paper as follows. \cref{sec:planning} presents the proposed cooperative path planning framework for the tethered robot duo. 
\cref{sec:control} describes the implementation of \ac{mrac} on the robot duo for robust trajectory tracking. \cref{sec:sim} and \cref{sec:experiment} show the simulation and experiment results with comprehensive evaluations, respectively. We conclude the paper in \cref{sec:conclusion}.

\section{Cooperative Path Planning}\label{sec:planning}

This section formally describes the proposed cooperative path planning framework for the tethered robot duo, assuming the environment and object locations are known.

\subsection{Problem Setup}

\begin{figure}[t!]
    \centering
    \includegraphics[width=\linewidth,trim=1cm 1cm 0cm 0cm, clip]{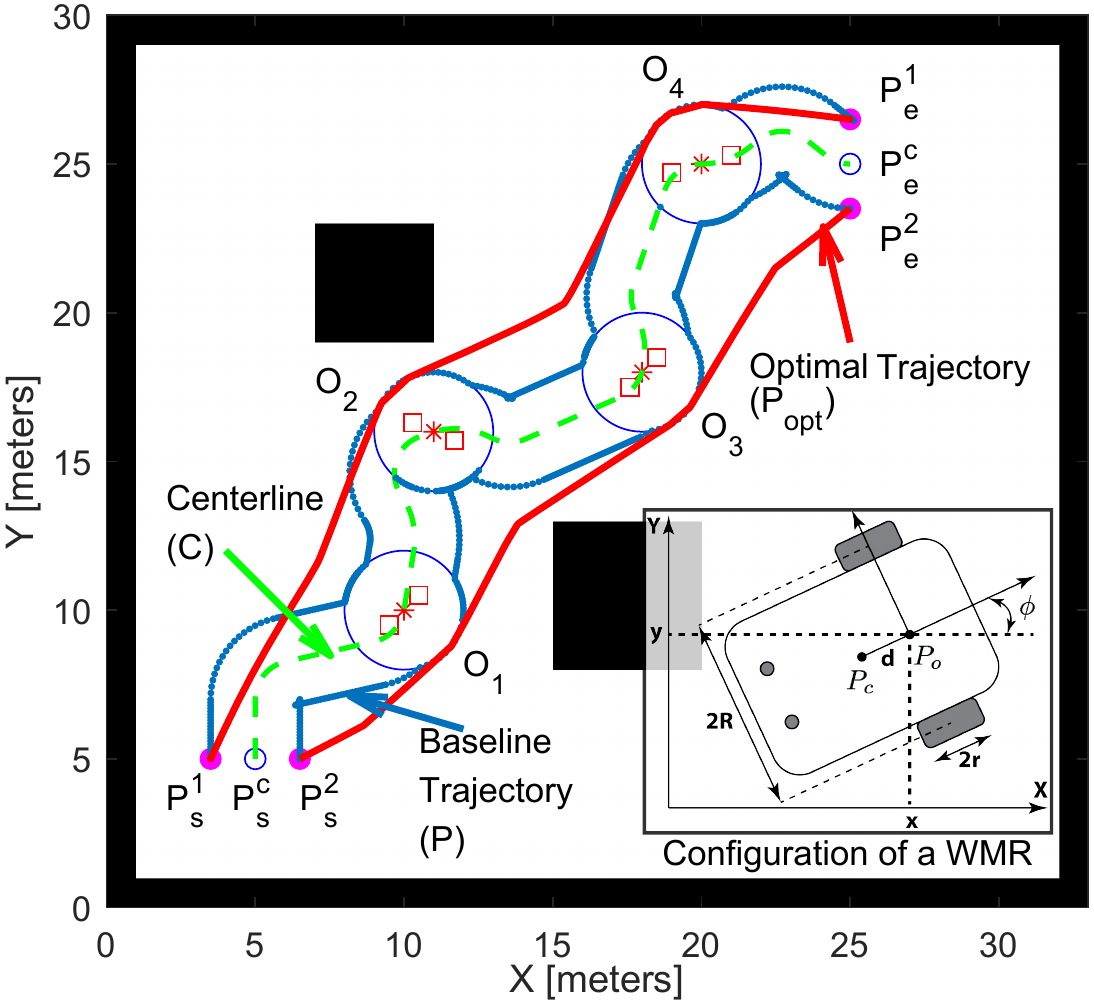}
    \caption{\textbf{The proposed cooperative path planning framework for tethered robot duo.} Our framework generates a \textcolor{green}{centerline} $C$ to connect start point $P^c_s$, end point $P^c_e$, and all object points, together with a set of waypoints (\ie, \textcolor{blue}{baseline trajectory} $P$), who serve as the initialization for the subsequent \textcolor{red}{trajectory optimization} to find $P_{opt}$.} 
    \label{fig:planning}
\end{figure}

\cref{fig:planning} illustrates our problem setup. Given an environment with a set of objects (red squares) and obstacles (black blocks), a robot duo is tasked to navigate from the initial configuration $P_s=(P^1_{s},P^2_{s})$ to reach the end configuration $P_e=(P^1_{e},P^2_{e})$ while collecting the objects on-the-way and avoiding obstacles. The environment is described as an occupancy grid $Map$, wherein $Map(i,j)=0$, $1$, and $2$ denote empty grids, grids occupied by obstacles, and grids occupied by objects, respectively. A safety margin $\gamma$ defined based on the robot's dimension is further added to obstacles for collision avoidance and to the radius of each target object's minimum bounding circle for safety. We also merge the circles that are closely overlapped. As a result, we have $M$ circles in total $O=\{O_1,\dots,O_M\}$, where $O_i=[x^o_i, y^o_i, r^o_i]$ includes the center position and radius of each circle. Of note, avoiding obstacles not only requires the robot duo to avoid directly colliding with the obstacle but further demands the tethered net not to enclose any obstacle at any moment. 

We devise a two-stage cooperative path planning framework to generate an optimal trajectory, shown also in \cref{fig:planning}. First, our framework generates a set of waypoints (blue dots) from the centerline path (dashed green line) that connects the robots' start and end configurations along with the circles that enclose target objects. We call these waypoints \textit{baseline trajectory}; they serve as a good initialization for the subsequent step. Next, our framework performs \textit{trajectory optimization} to produce feasible trajectories (solid red line).

\paragraph*{Assumption} Our framework has two assumptions: (i) The circles do not overlap with obstacles; namely, the objects cannot be too close to the obstacle such that the robot duo fails to navigate without violating the safety margin. (ii) The size of the circles is smaller than the length of the tethered net; the net constrains the robot duo's motions.

\subsection{Baseline Trajectory}\label{sec:plan_baseline}
 
Let the middle point of the tethered robot duo at the start configuration be $P^c_s=\frac{1}{2}(P^1_{s}+P^2_{s})$ and at the end configuration be $P^c_e=\frac{1}{2}(P^1_{e}+P^2_{e})$. First, we adopt a conventional path planning algorithm (hybrid $A^*$~\cite{petereit2012application} or $RRT^*$~\cite{gammell2014informed}) to construct the centerline by connecting $P^c_s$, each circle's center point $O_i$, and $P^c_e$; see dashed green line in \cref{fig:planning}. Formally, the centerline is denoted as $C=\{P^c_s,\dots,P^c_e\}\in\mathbb{R}^{N\times3}$, $P^c_i=[x^c_i,y^c_i,\phi^c_i]$, where $N$ is the total number of points. Next, we expand this centerline to form robot duo's baseline trajectories, $P=[P^l, P^r]\in\mathbb{R}^{N\times6}$; see blue curves in \cref{fig:planning}.

\begin{figure}[t!]
    \centering
    \begin{subfigure}[b]{0.27\linewidth}
        \centering
        \includegraphics[width=\linewidth]{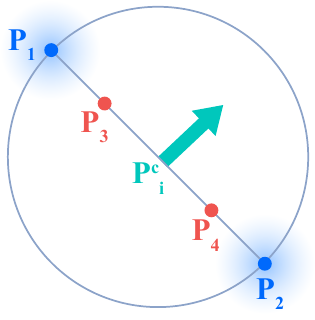}
        \caption{Case 1}
    \end{subfigure}%
    \begin{subfigure}[b]{0.38\linewidth}
        \centering
        \includegraphics[width=\linewidth]{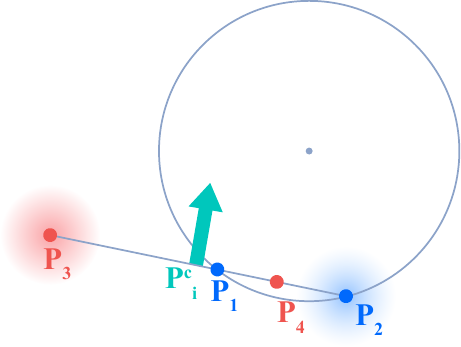}
        \caption{Case 2}
    \end{subfigure}%
    \begin{subfigure}[b]{0.35\linewidth}
        \centering
        \includegraphics[width=\linewidth]{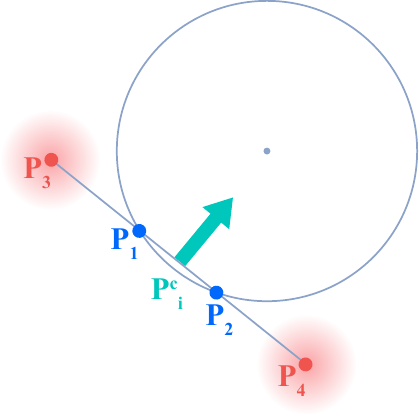}
        \caption{Case 3}
    \end{subfigure}%
    \caption{\textbf{The selection of baseline points.} Given the centerline point $P^c_i$ (green arrow), four candidates of baseline points are proposed; \cref{eq:line,eq:circle} yield $P_1$, $P_2$ (blue dots), and \cref{eq:vert} yields $P_3$, $P_4$ (red dots). Shaded points indicate the selected baseline points $P^l_{i}, P^r_{i}$.}
    \label{fig:baseline_select}
\end{figure}

Specifically, we divide $C$ into two groups: on or near the circumference of circles $C_{\text{circle}}$ and others $C_{\text{free}}$. For point $P^c_i\in{C_{\text{free}}}$, the baseline trajectory points are given by
\begin{equation}
    \small
    P^l_{i},P^r_{i} = \begin{bmatrix}
    x^c_{i}\,{\pm}\,l\cos(\phi^c_{i}+\frac{\pi}{2})\\
    y^c_{i}\,{\pm}\,l\sin(\phi^c_{i}+\frac{\pi}{2})\\
    \phi^{c}_i
    \end{bmatrix}^T,
    \label{eq:vert}
\end{equation}
where $l$ is the separation distance chosen based on the dimension of the \acp{wmr}.
For point $P^c_{i}\in{C_{\text{circle}}}$, we first calculate the line in the direction perpendicular to its orientation $\phi^c_i$ with
\begin{equation}
    \small
    y=kx+b,
    \label{eq:line}
\end{equation}
where $k=tan(\phi^c_{i}+\frac{\pi}{2})$, and $b=y^c_{i}-kx^c_{i}$. Next, we solve two intersection points, $P_1$ and $P_2$, by combining \cref{eq:line} with the related circle equation of $O_j$, 
\begin{equation}
    \small
    (x-x^o_{j})^2+(y-y^o_{j})^2={r^o_{j}}^2.
    \label{eq:circle}
\end{equation}
Another two points obtained by \cref{eq:vert} are denoted as $P_3$ and $P_4$. We choose $P^l_{i}, P^r_{i}$ from these four candidates by selecting the two outer ones, as shown in \cref{fig:baseline_select}.

\subsection{Trajectory Optimization}

With the estimated baseline trajectory in \cref{sec:plan_baseline} as the initialization, we further devise an iterative optimization framework to find the optimal trajectory for the tethered duo. Compared with optimizing the whole trajectory at once, our iterative optimization framework improves computational efficiency and avoids local minimum~\cite{byrne2014lecture,kelley1999iterative}. At each iteration, the trajectory of one mobile robot, denoted as $P^1$, is optimized, whereas the trajectory of the other robot, denoted as $P^2$, is treated as a known constant. \cref{alg:traj_opt} summarizes this framework, where $opt$ stands for the nonlinear optimization solver with defined cost functions and constraints.

\subsubsection{Cost Functions}
 
\paragraph*{Obstacle Cost}

We define obstacle cost in two forms: the cost of physical obstacles and the cost of objects as obstacles. The cost of physical obstacles is defined as
\begin{equation}
    J_{\text{obs},1}(i)=
    \left\{
        \begin{aligned}
            K_1, \qquad& Map(P^1(i,1),P^1(i,2))\geq1 \\
            0,\qquad  & Else
        \end{aligned}
    \right.,
\end{equation}
where $K_1$ is a positive parameter to be selected. 

The cost of objects as obstacles treats circles $O_j$ as obstacles, designed similarly to potential field methods~\cite{lee2003artificial}:
\begin{equation}
    J_{\text{obs},2}(i,j)=
    \left\{
        \begin{aligned}
            \frac{1}{2}K_2(\frac{1}{D(i,j)}-\frac{1}{r^o_{j}})^2, \quad & D(i,j){\leq}r^o_{j} \\
            0,\quad &D(i,j) > r^o_{j} 
        \end{aligned}
    \right.,
\end{equation}
where $D(i,j)$ denotes the Euclidean distance between the trajectory point $P^1_{i}$ and the object circle center $O_j$, and $K_2$ is a positive parameter to penalize the distance.

As such, the combined obstacle cost of $P^1_{i}$ is defined as
\begin{equation}
    J_{\text{obs}}(i)=J_{\text{obs},1}(i)+\sum_{j=1}^{M} J_{\text{obs},2}(i,j).
\end{equation}

\paragraph*{Distance Cost}

The distance cost is calculated as the weighted sum of the Euclidean distance between two consecutive points and the orientation difference:
\begin{equation}
    \begin{aligned}
        J_{\text{dist}}(i)&=t_1\sqrt{\Delta x_i^2+\Delta y_i^2}+t_2\lvert\Delta \phi_i\rvert,\\
        \Delta x_i&=P^1(i,1)-P^1(i+1,1),\\
        \Delta y_i&=P^1(i,2)-P^1(i+1,2),\\
        \Delta \phi_i&=P^1(i,3)-P^1(i+1,3),
    \end{aligned}
\end{equation}
where $t_1$ and $t_2$ are two non-negative parameters.

\paragraph*{Expansion Cost}

The cost function for expansion is designed as a U-shape cost function~\cite{ryll2015}:
\begin{equation}
    \resizebox{0.91\linewidth}{!}{%
        $\displaystyle{}J_{e}(i)=
        \begin{cases} 
            k_{d1}\tan^2(\gamma_1d+\gamma_2), & d_{\text{min}} \leq d \leq d_{\text{rest}} \\
            k_{d2}(d-d_{\text{rest}})^2-\frac{(d-d_{\text{rest}})^2}{(d-d_{\text{max}})^2}, & d_{\text{rest}} <  d \leq d_{\text{max}} \\
            K_3,  &d < d_{\text{min}} \ \text{or} \ d > d_{\text{max}}
        \end{cases}$%
    }%
    \label{eq:expan_cost}
\end{equation}
where $\gamma_1=\frac{\pi}{2(d_{\text{rest}}-d_{\text{min}})}$, and $\gamma_2=-\gamma_1d_{\text{rest}}$. $d$ is the Euclidean distance between trajectory points $P^1_{i}$ and $P^2_{i}$, representing the expansion of the tethered net. $k_{d1}$, $k_{d2}$, and $K_3$ are gain parameters. $d_{\text{max}}$, $d_{\text{min}}$, and $d_{\text{rest}}$ denote the minimum, maximum, and the ideal expansion of the tethered net, set based on various hardware setup and task scenarios.

The reason for having this expansion cost is to maintain the U-shape of the tethered net. Conversely, the net shape would make it infeasible to collect objects when the two \acp{wmr} are too close or too far. In this paper, we set $d_{\text{min}}=0.1d_{\text{max}}$ and $d_{\text{rest}}=\frac{2}{\pi}d_{\text{max}}$\textemdash{}a half-circle shape. Of note, if $d_{\text{max}}$ is not specified, we can treat it as a random variable to estimate the optimal net length in certain task scenarios.

\paragraph*{Smoothness Cost}

The smoothness cost is defined as the sum of squared accelerations along the trajectory:
\begin{equation}
    J_s=\sum_{j=1}^{3} b_j P^1(:,j)^T QP^1(:,j),
\end{equation}
where $Q=A^TA$, and $A$ is the finite difference matrix~\cite{kalakrishnan2011stomp}, $b_j$s are weighting gains.

Taken together, the total cost along the whole trajectory is defined by the sum of the above cost functions:
\begin{equation}
    \resizebox{0.89\linewidth}{!}{%
        $\displaystyle{}J_{\text{total}}=a_1J_s+\sum_{i=1}^{N}a_2J_{\text{dist}}(i)+ a_3J_{\text{obs}}(i)+a_4J_e(i)$
    }%
\end{equation}
where $a_1, a_2, a_3,$ and $a_4$ are weighting gains for costs.

\subsubsection{Constraints}

\begin{algorithm}[t!]
    \small
    \caption{Trajectory Optimization Algorithm}
    \KwData 
            {$N_{\text{total}}, P, O, Map, t_{1-2},K_{1-3},k_{d1},k_{d2}$\\ 
             $\qquad\quad Q,b_{1-3},a_{1-4},\Delta L_{\text{max}}, \Delta\phi_{\text{max}}$ }
    \KwResult {$P_{opt}, d_{\text{max}}$} 
        $step\gets1, d_{\text{max}}\gets N\Delta L_{\text{max}}$\tcp*{Initialization}
        \While{$step\leq N_{\text{total}}$}{  
          \eIf(\tcp*[h]{Optimize left half}){$mod(step,2)=1$}{ 
             $P_s\gets P(1,1:3), P_e\gets P(N,1:3)$\;
             \tcp*[h]{Specify start and end points}
             $P^1_{ini}\gets P(2:N\text{-1},1:3)$\;
             \tcp*[h]{Traj to be optimized}
             $P^2_{ini}\gets P(2:N\text{-1},4:6)$\;
             \tcp*[h]{Traj of another half as constant}
            \resizebox{0.8\linewidth}{!}{$[P^1_{opt},d_{\text{max}}]\gets opt(P^1_{ini},P^2_{ini},P_s,P_e,Map,O,d_{\text{max}})$\; }
            \tcp*[h]{Optimization process} 
             $P(2:N\text{-1},1:3)\gets P^1_{opt}$\;
             \tcp*[h]{Update left half trajectory}
              }
          (\tcp*[h]{Optimize right half}){    
            $P_s\gets P(1,4:6), P_e\gets P(N,4:6)$ \;   
            $P^1_{ini}\gets P(2:N\text{-1},4:6)$\;
            $P^2_{ini}\gets P(2:N\text{-1},1:3)$\;
            \resizebox{0.8\linewidth}{!}{    $[P^2_{opt},d_{\text{max}}]\gets opt(P^1_{ini},P^2_{ini},P_s,P_e,Map,O,d_{\text{max}})$\;}
            $P(2:N\text{-1},4:6)\gets P^2_{opt}$\;
          }
           $step\gets step+1$\tcp*{Update steps}
        }
    $P_{opt}\gets P$ \tcp*{Output optimized trajectory}
    \label{alg:traj_opt}
\end{algorithm}

\paragraph*{Maximum Velocity Constraint}

We define it as:
\begin{equation}
    \small
    \begin{aligned}
        0 \leq \sqrt{\Delta x_i^2+\Delta y_i^2} \leq \Delta L_{\text{max}},&\\
        -\Delta \phi_{\text{max}} \leq \Delta \phi_i \leq \Delta \phi_{\text{max}},&
    \end{aligned}
\end{equation}
where $\Delta L_{\text{max}}$ is the maximum travel distance at each step, and $\Delta \phi_{\text{max}}$ is the maximum turn angle at each step.

\paragraph*{Object and Obstacle Constraint}

We select consecutive points on trajectory $P^1$ and $P^2$ (\ie, $P^1_{i}, P^1_{i+1}, P^2_{i}, P^2_{i+1}$) to build a quadrilateral, resulting in $N-1$ quadrilaterals. To ensure collecting all the objects, the centers of all circles $O$ must be covered by these quadrilaterals. Moreover, to prevent the path from enclosing any obstacles inside, these quadrilaterals must not overlap with any obstacles.

\section{Robust Trajectory Tracking Control}\label{sec:control}

We adopt a decentralized framework to design the trajectory tracking controller for individual \acp{wmr}. Since the drag force from the tethered net and gathered objects are unknown in our proposed setup, the conventional model-based controller is not suitable, and a more advanced \ac{mrac} is implemented to improve the robustness of the trajectory tracking control.

\subsection{Dynamics of Individual \texorpdfstring{\acp{wmr}}{}}

The configuration of \acp{wmr} is defined by $q=[x, y, \phi]$, shown also in \cref{fig:planning}. Li \etal~\cite{li2020trajectory} describe its dynamics:
\begin{equation}
    \small
    M(q)\Ddot{q}+C(q,\Dot{q})\Dot{q}=B(q)\tau-A(q)^T\lambda,
    \label{eq:dyn eq}
\end{equation}
where
\begin{equation}
    \small
    M(q)=\begin{bmatrix}
        m& 0 & {m}d\sin\phi\\
        0& m & -{m}d\cos\phi\\
        {m}d\sin\phi& -{m}d\cos\phi & J
    \end{bmatrix},
\end{equation}
\begin{equation}
    \small
    C(q,\Dot{q})=\begin{bmatrix}
        0& 0 & {m}d\Dot{\phi}\cos\phi\\
        0& 0 & -{m}d\Dot{\phi}\sin\phi\\
        0& 0 & 0
    \end{bmatrix},
\end{equation}
\begin{equation}
    \small
    B(q)=\frac{1}{r}
    \begin{bmatrix}
        \cos\phi& \cos\phi\\
        \sin\phi& \sin\phi\\
        R&   -R
    \end{bmatrix},
\end{equation}
\begin{equation}
    \small
    A(q)=\begin{bmatrix}
        -\sin\phi&\cos\phi&-d 
    \end{bmatrix},
\end{equation}
\begin{equation}
    \small
    \lambda=-m(\Dot{x}\cos\phi+\Dot{y}\sin\phi)\Dot{\phi},
\end{equation}
where $m$ is the \ac{wmr}'s mass, $J$ its rotational inertia, $2R$ the distance between two driving wheels, $r$ the radius of each wheel, $d$ the distance from the coordinate origin $P_0$ to the CoM $P_c$, $M(q)$ the inertia matrix, $C(q,\Dot{q})$ the Coriolis and Centrifugal force matrix, $B(q)$ the input transformation matrix, $\tau$ the torque vector for two wheels, $A(q)$ matrix associated with the non-holonomic constraints (\ie, $A(q)\Dot{q}=0$), and $\lambda$ the vector of constraint forces.
Selecting $S(q)$ as a basis of $A(q)$ nullspace (\ie, $S(q)^TA(q)^T=0$), 
\begin{equation}
    \small
    S(q)=\begin{bmatrix}
    \cos\phi &-d\sin\phi\\
    \sin\phi &d\cos\phi\\
    0 &1
    \end{bmatrix},
\end{equation}
we can rewrite the \ac{wmr}'s kinematic equation with the velocity vector,
\begin{equation}
    \small
    \Dot{q}=S(q)v,
    \label{eq:kinematic}
\end{equation}
where $v=[\nu,\omega]^T$, and $\nu$ and $\omega$ are the \acp{wmr}' linear and angular velocity. Taking the derivative of \cref{eq:kinematic}, we have
\begin{equation}
    \small
    \Ddot{q}=\Dot{S}(q)v+S(q)\Dot{v}.
    \label{eq:kinematic_dot}
\end{equation}
Substituting \cref{eq:kinematic,eq:kinematic_dot} into \cref{eq:dyn eq} and multiplying $S^T$ at each side, we have
\begin{equation}
    \small
    \Bar{M}\Dot{v}+\Bar{C}v=\Bar{\tau},
    \label{eq:dyn_eq_final}
\end{equation}
where $\Bar{M}=S(q)^TM(q)S(q)$, $\Bar{\tau}=\Bar{B}\tau$, $\Bar{B}=S(q)^TB(q)$, and $\Bar{C}=S(q)M(q)\Dot{S}(q)+S(q)^TC(q,\Dot{q})S(q)$.

\subsection{Model-Based Control of Individual \texorpdfstring{\acp{wmr}}{}}

We design a conventional model-based controller based on \cref{eq:dyn_eq_final} as a baseline,
\begin{equation}
    \small
    \tau=\Bar{B}^\dagger(\Bar{M}\Dot{v}^d+\Bar{C}v^d),
\end{equation}
where $v^d$ is the desired velocity vector, designed to track the reference trajectory~\cite{li2020trajectory}:
\begin{equation}
    \resizebox{0.89\linewidth}{!}{%
        $v^d=\begin{bmatrix}
            \nu^d\\ \omega^d
        \end{bmatrix}
        =
        \begin{bmatrix}
            {\nu^r}{\cos}e_{\phi}+k_1(e_x+d(1-{\cos}e_{\phi}))
            \\ \omega^r+k_2\nu^r(e_y-d{\sin}e_{\phi}+k_3\nu^r{\sin}e_{\phi})
        \end{bmatrix},$
    }%
\end{equation}
where $\nu^r$ and $\omega^r$ are reference linear and angular velocity, $k_1$, $k_2$, and $k_3$ are positive parameters, and $e_x$, $e_y$, and $e_{\phi}$ are errors between the reference trajectory and the real trajectory in $x$, $y$, and $\phi$, respectively~\cite{li2020trajectory},
\begin{equation}
    \small
    e_x = x^r - x,\,
    e_y = y^r - y,\,
    e_\phi = \phi^r - \phi.
\end{equation}%
Although this controller has proven to work well on \acp{wmr} given accurate physical parameters, it fails in our setting because the tethered net will collect multiple objects with unknown physic parameters (mass, inertia, \etc). As such, we further design an adaptive control law; see the next section.

\subsection{Adaptive Control with Unknown Payload}

Since \ac{mrac} laws guarantee the asymptotic convergence of trajectory tracking error in the presence of parametric and matched uncertainties~\cite{lavretsky2013robust}, we implement \ac{mrac} on \acp{wmr} to improve its robustness under unknown payload.

Specifically, we rewrite the matrices in \cref{eq:dyn_eq_final} with unknown dynamics,
\begin{equation}
    \begin{aligned}
        \Bar{M}&=\Bar{M}_k+\Bar{M}_u^*=(\Bar{M}_k-I)+(I+\Bar{M}_u^*)\\
        &=(\Bar{M}_k-I)+\Bar{M}_u,\\
        \Bar{C}&=\Bar{C}_k+\Bar{C}_u,
    \end{aligned}
    \label{eq:add_unknown}
\end{equation}
where $k$ and $u$ denote the known and unknown dynamics, respectively. Combining 
\cref{eq:add_unknown} with \cref{eq:dyn_eq_final}, we have 
\begin{equation}
    \Dot{X}(t)=A_{r}X(t)+B_{r}u(t),
\end{equation}
where $X=v$, $A_{r}=-\Bar{M}_u^{-1}\Bar{C}_u$, and $B_{r}=\Bar{M}_u^{-1}$.
The input $\Bar{\tau}$ can be recovered by
\begin{equation}
    \Bar{\tau}=u+(\Bar{M}_k-I)\Dot{v}(t)+\Bar{C}_kv(t),
\end{equation}
and the reference model is given as
\begin{equation}
    \Dot{X}_{m}(t)=A_{m}X_{m}(t)+B_{m}u_m(t),
\end{equation}
where $X_{m}=v^d$, $A_m=-K$, $K$ is a Hurwitz matrix, and $B_m=I$.
The reference input $u_m$ can be calculated by
\begin{equation}
    u_m=B_m^{-1}(\Dot{X}_m-A_mX_m)
       =\Dot{v}^d+Kv^d.
\end{equation}
The adaptive law is designed as
\begin{equation}
    \begin{aligned}
        u&=\Bar{u}_a+\Bar{u}_k,\\
        \Bar{u}_k&=-KX+u_m,\\
        \Bar{u}_a&=-\Delta_XX+\Delta_mu_m, 
    \end{aligned}
\end{equation}
where $\Bar{u}_k$ and $\Bar{u}_a$ are the linear and adaptive feedback component, respectively. $\dot{\Delta}_X$ and $\dot{\Delta}_m$ can be calculated with adaptive gains and state errors; please refer to Canigur \etal~\cite{canigur2012model} for more details. Of note, the motor torque saturation and the adaptive gains of \ac{mrac} jointly determine the maximum payload while the stability of tracking controller is still maintained. Larger adaptive gains can increase the response speed but may lead to oscillations and higher overshoot~\cite{stellet2011influence}.

\section{Simulation}\label{sec:sim}

This section evaluates our cooperative planning framework in a simulation environment based on Gazebo, a 3D dynamic environment simulator. Our results demonstrate that (i) the proposed framework produces optimal trajectories for the tethered robots, and (ii) our U-shape cost function design effectively maintains the tethered net during the tasks.

\subsection{Setup}

\begin{figure}[t!]
    \centering
    \includegraphics[width=\linewidth]{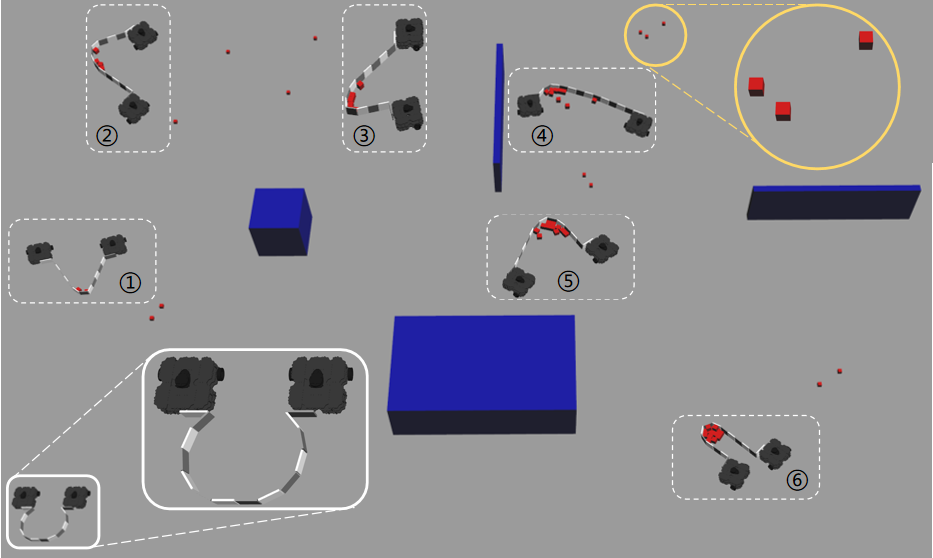}
    \caption{\textbf{The simulation environment.} The tethered robot duo is tasked to gather 13 objects (red cubes), navigating from the bottom left to the bottom right. The dashed boxes depict some key moments when the duo gathers objects with the flexible net.}
    \label{fig:sim_setup}
\end{figure}

\begin{figure*}[t!]
    \centering    
    \begin{subfigure}[b]{0.333\linewidth}
        \centering
        \includegraphics[width=\linewidth,trim=4.3cm 10.5cm 4cm 10.2cm,clip]{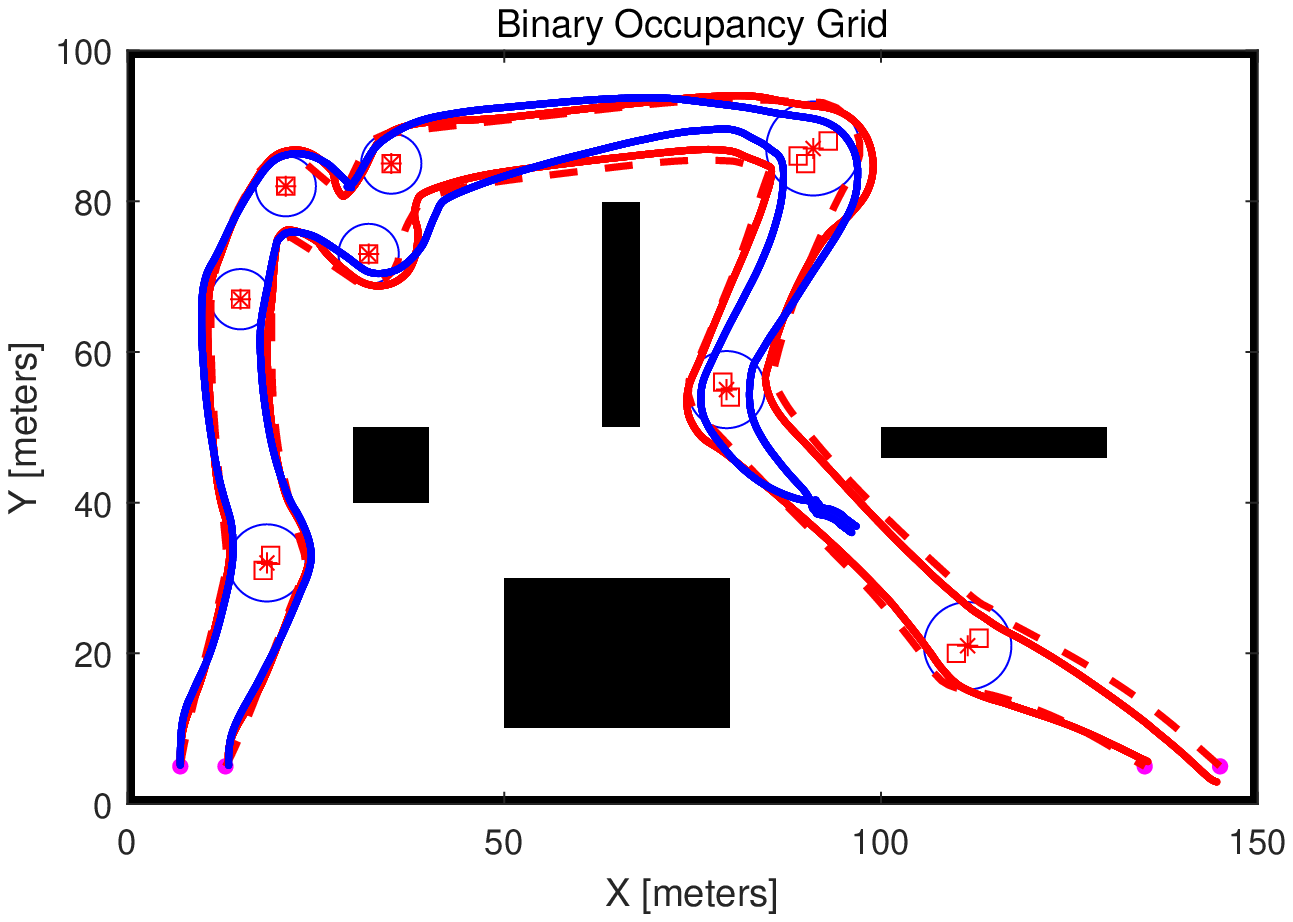}
        \caption{Trajectory tracking $d_{\text{max}}=1.5~m$}
        \label{fig:sim_planning_15}
    \end{subfigure}%
    \begin{subfigure}[b]{0.333\linewidth}
        \centering
        \includegraphics[width=\linewidth,trim=4.3cm 10.5cm 4cm 10.2cm, clip]{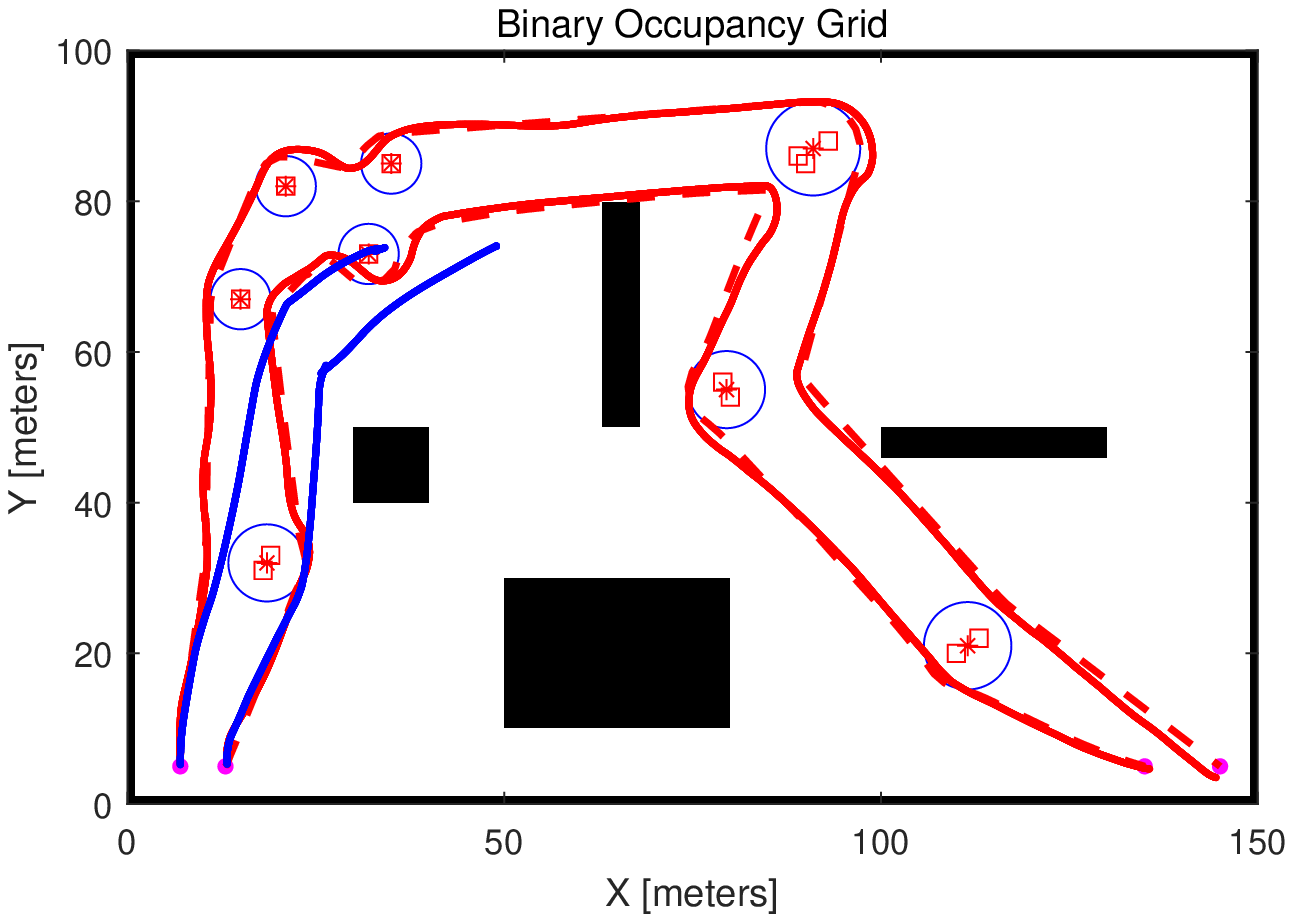}
        \caption{Trajectory tracking $d_{\text{max}}=2.0~m$}
        \label{fig:sim_planning_20}
    \end{subfigure}%
    \begin{subfigure}[b]{0.333\linewidth}
        \centering
        \includegraphics[width=\linewidth,trim=4.3cm 10.5cm 4cm 10.2cm, clip]{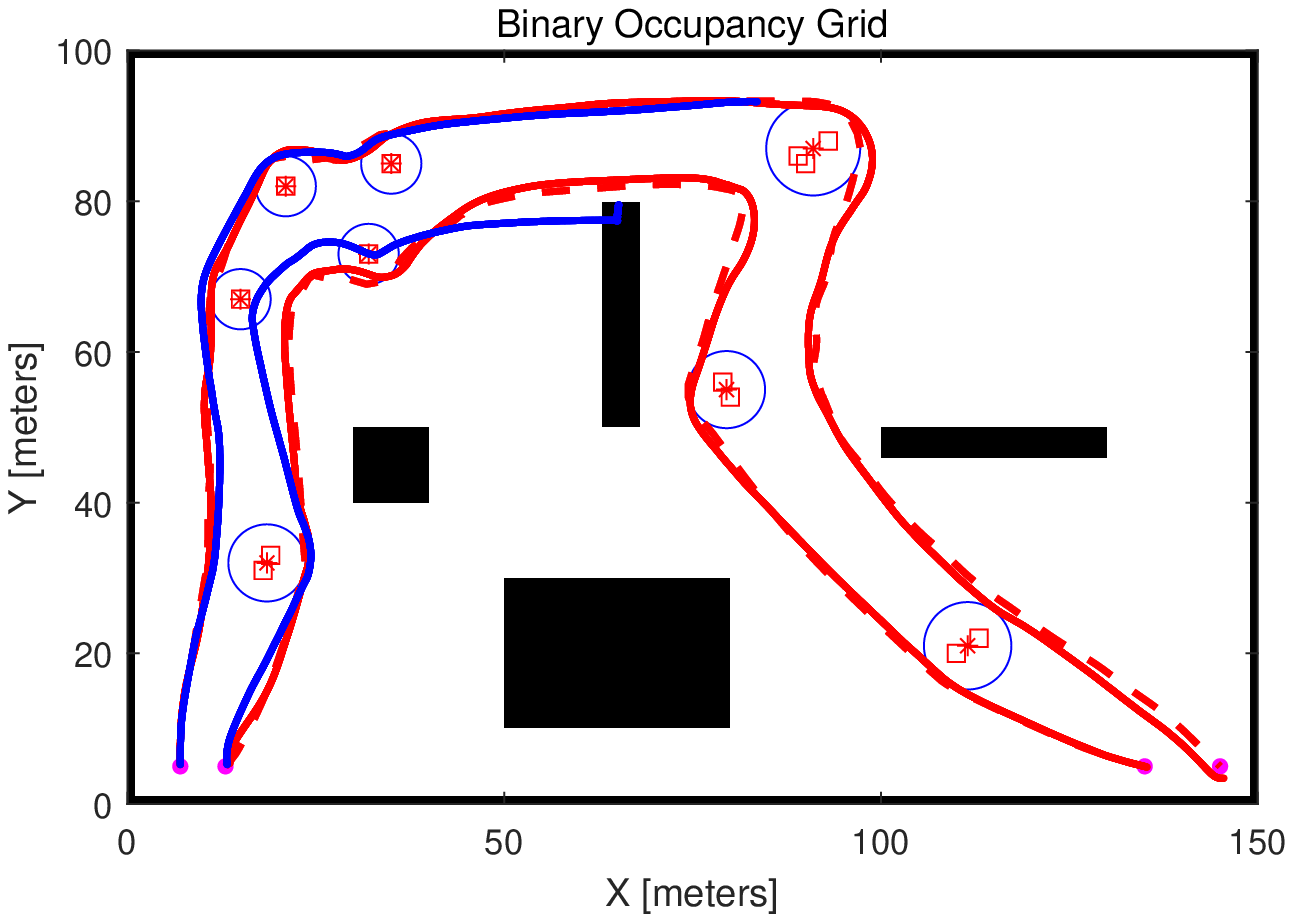}
        \caption{Trajectory tracking $d_{\text{max}}=2.32~m$}
        \label{fig:sim_planning_23}
    \end{subfigure}%
    \\%
    \begin{subfigure}[b]{0.333\linewidth}
        \centering
        \includegraphics[width=\linewidth,trim=4.3cm 10.5cm 4cm 10.2cm, clip]{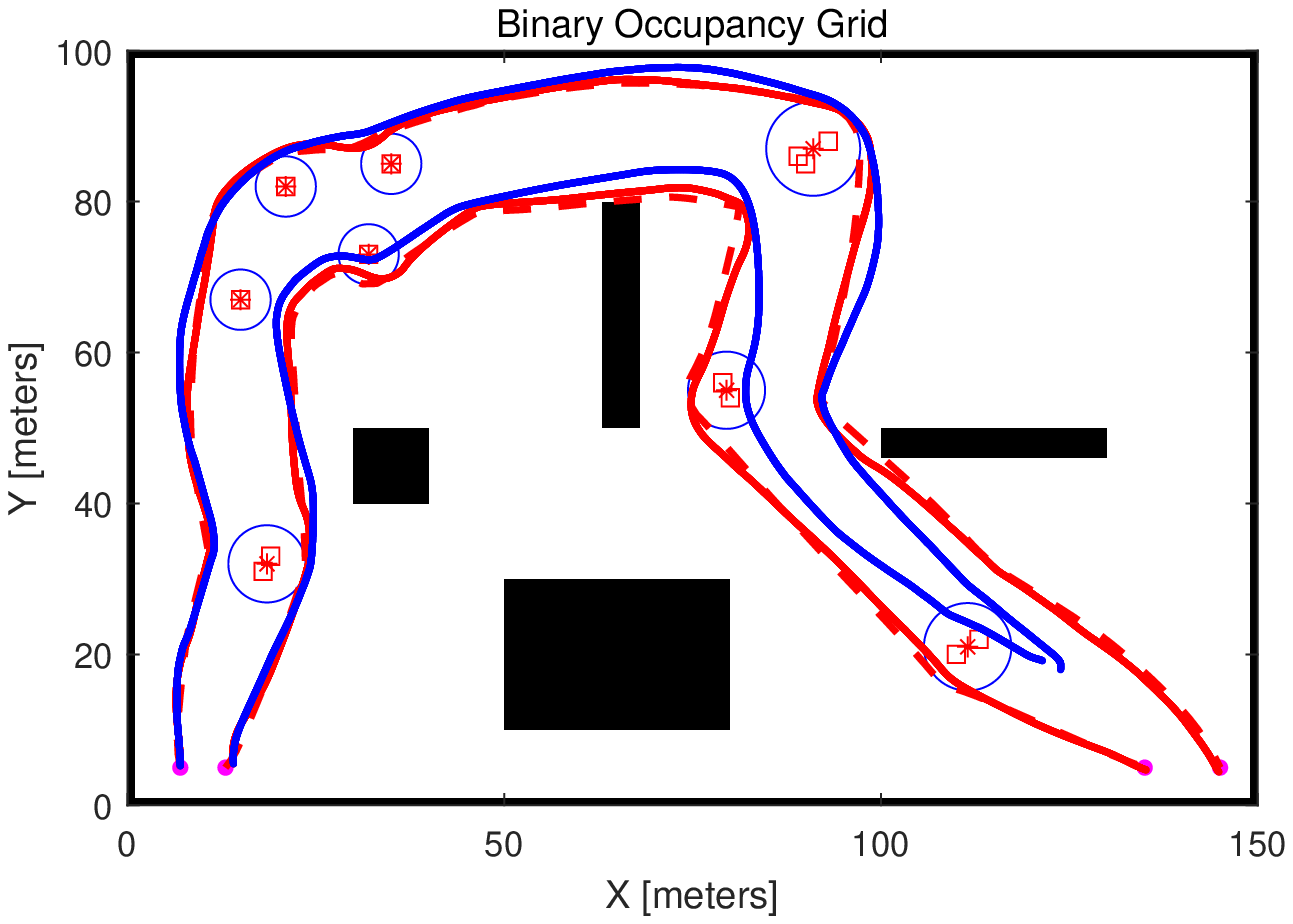}
        \caption{Trajectory tracking $d_{\text{max}}=2.5~m$}
        \label{fig:sim_planning_25}
    \end{subfigure}%
    \begin{subfigure}[b]{0.333\linewidth}
        \centering
        \includegraphics[width=\linewidth,trim=4.3cm 10.5cm 4cm 10.2cm, clip]{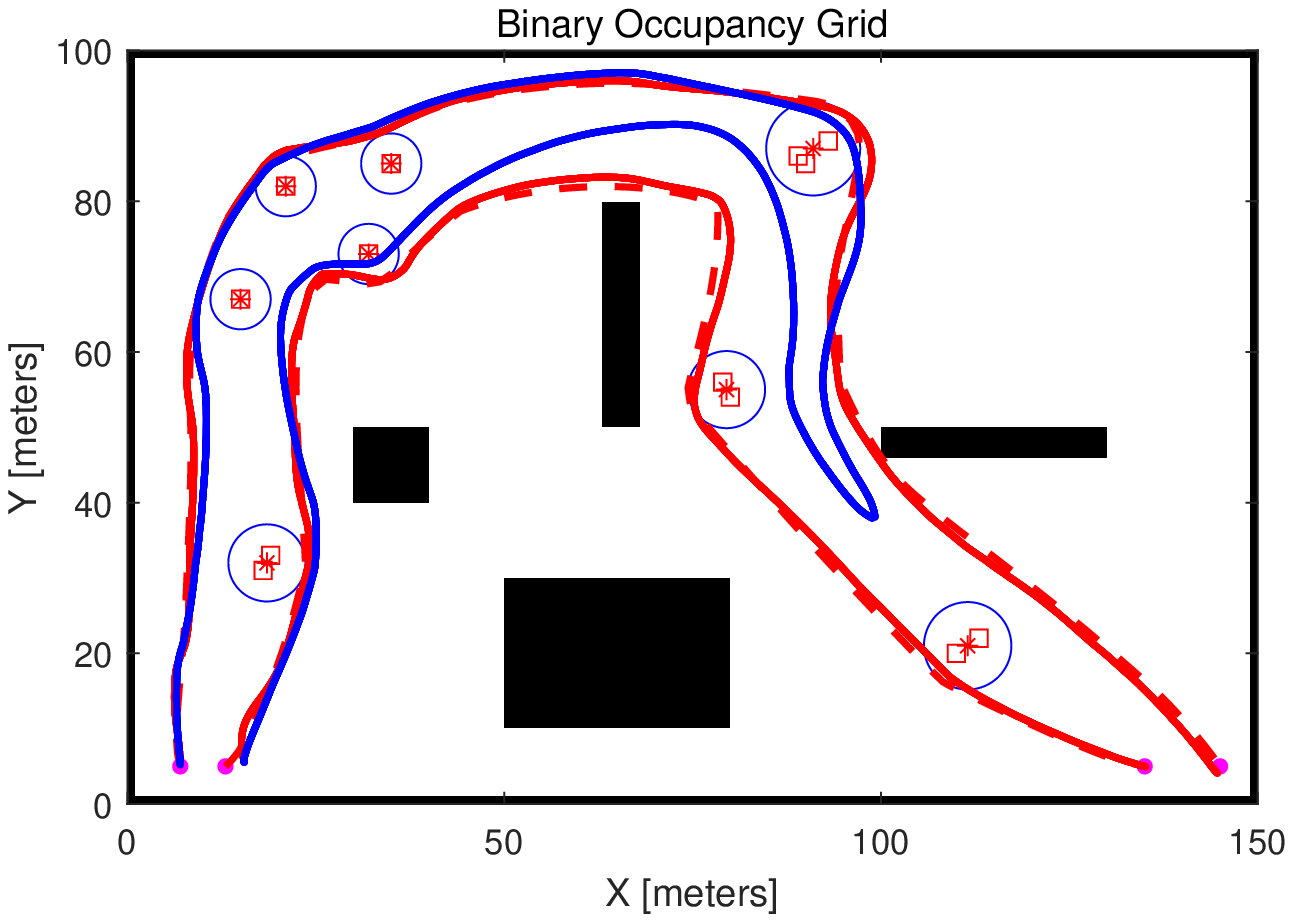}
        \caption{Trajectory tracking $d_{\text{max}}=3.0~m$}
        \label{fig:sim_planning_30}
    \end{subfigure}%
    \begin{subfigure}[b]{0.32\linewidth}
        \centering
        \includegraphics[width=\linewidth,trim=4.0cm 10.0cm 4cm 10cm,clip]{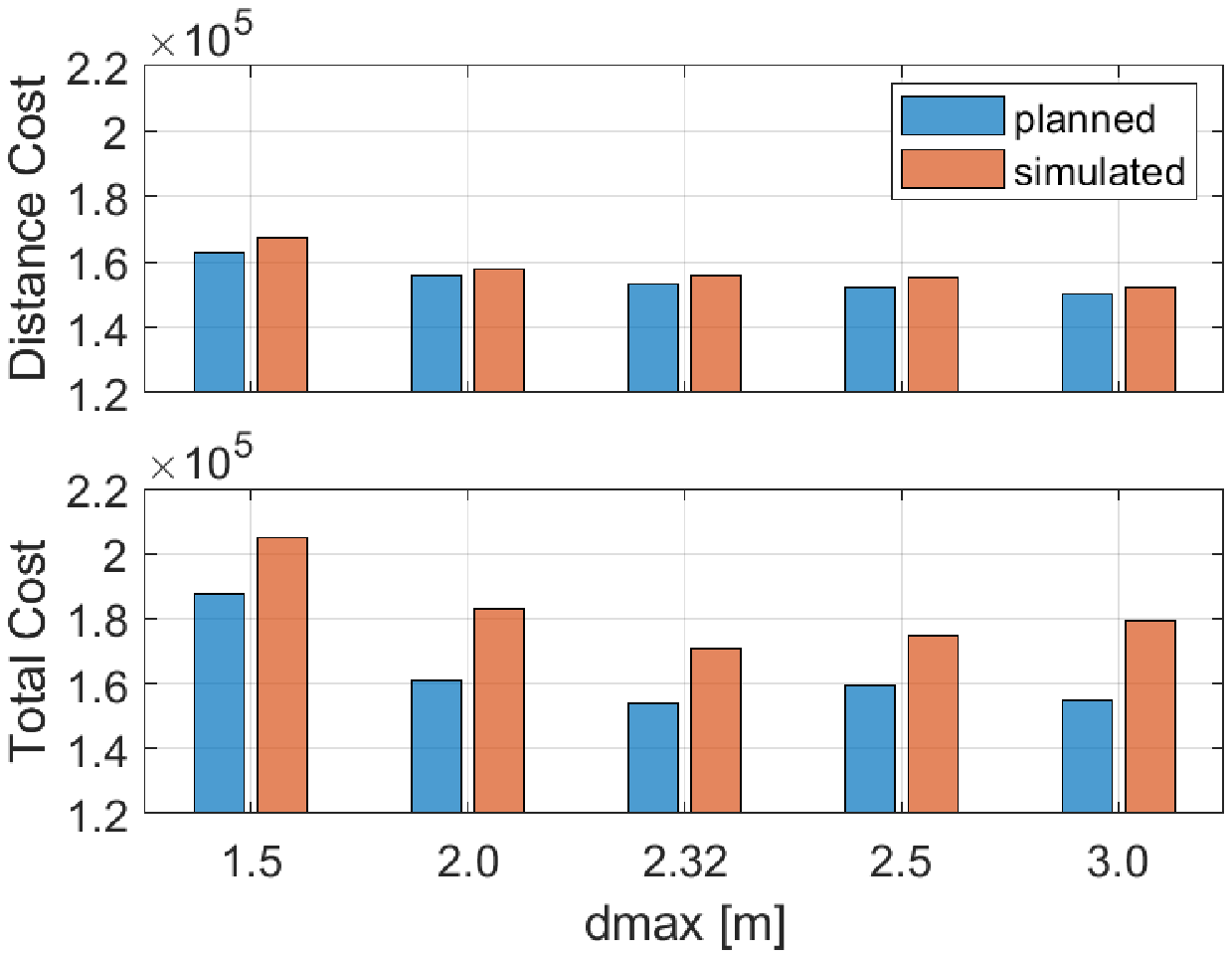}
        \caption{Simulation cost and planing cost}
        \label{fig:cost}
    \end{subfigure}%
    \caption{(a--e) \textbf{The schematic of the simulated environment constructed in Gazebo.} With various lengths of the tethered net, our cooperative planning framework produces reference trajectory (dashed red lines). Tracked trajectories by the proposed \ac{mrac} controller and conventional model-based controller are visualized in solid red and blue lines, respectively; the \ac{mrac} controller can robustly track trajectories, whereas the conventional controller constantly fails as the task progresses. (f) \textbf{Distance cost and total cost of the optimized trajectories in the planning and the simulated execution phases.} Our framework further produces the optimal $d_{\text{max}} = 2.32~m$ for a given configuration.} 
    \label{fig:sim_planning}
\end{figure*}

For the tethered robot platform, we utilize two TurtleBot3 Waffle Pi robots and tether them with a flexible ``net'' by connecting several 0.1-meter-long thin cuboid links with passive revolute joints; the length of the net can be easily modified by inserting or removing cuboid links. With such a design, the tethered net would deform in accord to gathered objects' weights, and drag forces could be introduced to robots in an unspecified direction, crucial in evaluating our \ac{mrac} implementation. To enable torque command to the robots in simulation, we use \emph{JointEffortController} in \emph{ros\_control} to set desired torques to TurtleBot3's wheels instead of using the built-in controller with velocity control only.

\cref{fig:sim_setup} illustrates the simulated environment with several keyframes of a task performance. The tethered robot duo starts from the bottom left to gather red movable cubes while avoiding blue obstacles at various locations. \cref{tab:setup} tabulates some essential physical and software properties of the simulation setup. We set the parameters involved in the trajectory planning as following:
$l=0.3~m$, $N_{\text{total}}=8$, $t_1=300$, $t_2=10$, $K_1=K_2=K_3=1e7$, $k_{d1}=20$, $k_{d2}=15$, $b_1=b_2=110$, $b_3=1$, $a_1=a_2=a_3=a_4=1$, $\Delta L_{\text{max}}=0.05~m$, and $\Delta \phi_{\text{max}}=0.1~rad$.
We utilize the nonlinear optimization solver \textsf{fmincon} in Matlab to solve the optimal trajectory with the interior point method.

\begin{table}[b!]
    \centering
    \caption{Physical and Software Properties used in Simulation}
    \label{tab:setup}
    \resizebox{\linewidth}{!}{%
        \begin{tabular}{c c c}
            \toprule
            \textbf{Group} &  \textbf{Parameter} &  \textbf{Value}\\
            \midrule
            \multirow{5}{*}{\rotatebox[origin=c]{90}{Robot}}
                & Mass $m$ & $1.43~kg$\\
                &  Moment $J$ & $0.146~kg \cdot m^2$\\
                & $R$ & $0.144~m$\\
                &  $r$ & $0.033~m$\\
                &  $d$ & $0.020~m$\\
            \midrule
            \multirow{3}{*}{\rotatebox[origin=c]{90}{Net}}
                & Mass of each cuboid link  &$2~g$\\
                & Size of each cuboid link $d\times w \times h$ & $1\times10\times8~ cm^3$\\
                & Damping of each cuboid joint  & $1e-3$\\
                &Friction of each cuboid joint  & $\mu_1=\mu_2=0.2$\\
            \midrule
            \multirow{4}{*}{\rotatebox[origin=c]{90}{Others}} 
                & Mass of object cube $m_i$ & $25~g$\\
                & Size of object cube  &$5\times5\times5~cm^3$\\
                & Trajectory tracking controller rate & $100~Hz$ \\
                & Motor torque controller rate & $500~Hz$\\
            \bottomrule
        \end{tabular}%
    }%
\end{table}

\subsection{Trajectory Tracking}

\cref{fig:sim_planning} shows the simulation results with various lengths of the tethered net $d_{\text{max}}$. Given a configuration during the optimization, our framework can further estimate the optimal length of the tethered net in terms of defined cost function $J_\textit{total}$ by treating $d_{\text{max}}$ as a variable in \cref{alg:traj_opt}. If $d_{\text{max}}$ is already specified, \cref{alg:traj_opt} can be easily modified by treating $d_{\text{max}}$ as a constant instead of a variable. Among five cases with various maximum net lengths shown in \cref{fig:sim_planning}, our framework estimates $d_{\text{max}}=2.32~m$ is the optimal net length.

In these simulations, we implement both \ac{mrac} controller and the conventional model-based controller. Specifically, the \ac{mrac} controller can maintain the platform stability along the whole trajectory while successfully gathering and transporting all the objects to the endpoint. By contrast, the conventional model-based controller is unstable when more objects are collected, resulting in collisions between the two robots or between the robot and objects/obstacles.

For each of the above cases, the distance cost $J_{\text{dist}}$ and total cost $J_{\text{total}}$ are first obtained with recorded robot trajectory. Next, these costs are compared with planned costs; the results are shown in \cref{fig:cost}. In particular, if we only care about $J_{\text{dist}}$, both $d_{\text{max}}=2.5~m$ and $d_{\text{max}}=3.0~m$ are better. However, it is difficult to maintain the shape of the tethered net under these two settings. As such, our optimization framework deems they are not the optimal solution. In comparison, although $d_{\text{max}}=2.32~m$ results in a slightly higher $J_{\text{dist}}$, this length is considered as the optimal one due to the lowest $J_{\text{total}}$.

This simulation result demonstrates the tracking performance of the proposed \ac{mrac} controller to handle increasing payloads. It also suggests that a longer tether between robots does not necessarily correspond to better efficiency. This finding may impact prior arts wherein the length is not considered~\cite{kim2013topological,bhattacharya2015topological} or is fixed~\cite{teshnizi2021motion,d2021catenary,talke2018catenary,laranjeira2017catenary,laranjeira2020catenary}.

\subsection{Maintaining Net Shape}

\begin{figure}[t!]
    \centering
    \begin{subfigure}[b]{0.5\linewidth}
        \centering
        \includegraphics[width=\linewidth,trim=4.4cm 10.5cm 4cm 10cm,clip]{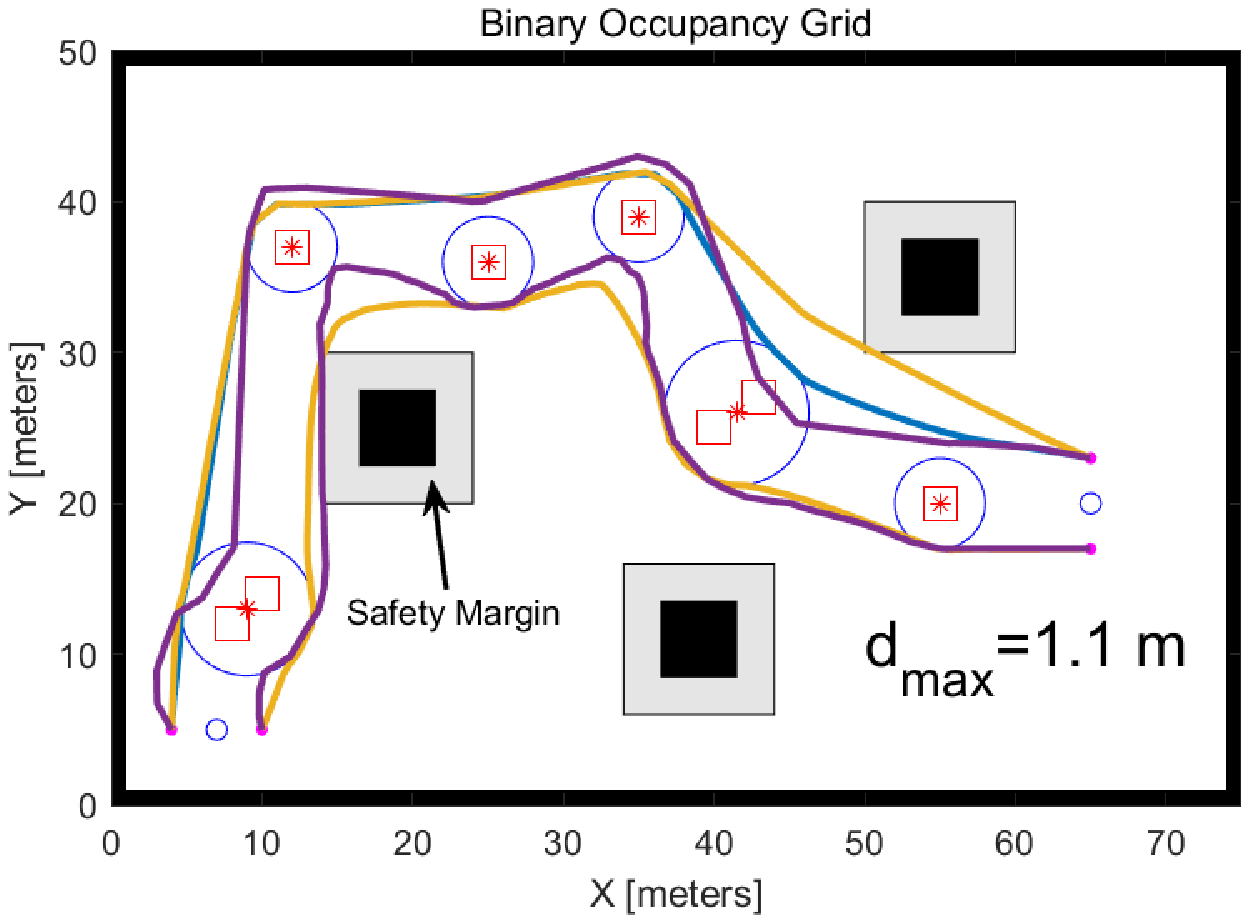}
        \caption{Trajectory comparison} 
        \label{fig:compare_shape_traj}
    \end{subfigure}%
    \begin{subfigure}[b]{0.5\linewidth}
        \centering
        \includegraphics[width=\linewidth,trim=4.0cm 10.3cm 3.8cm 10.4cm,clip]{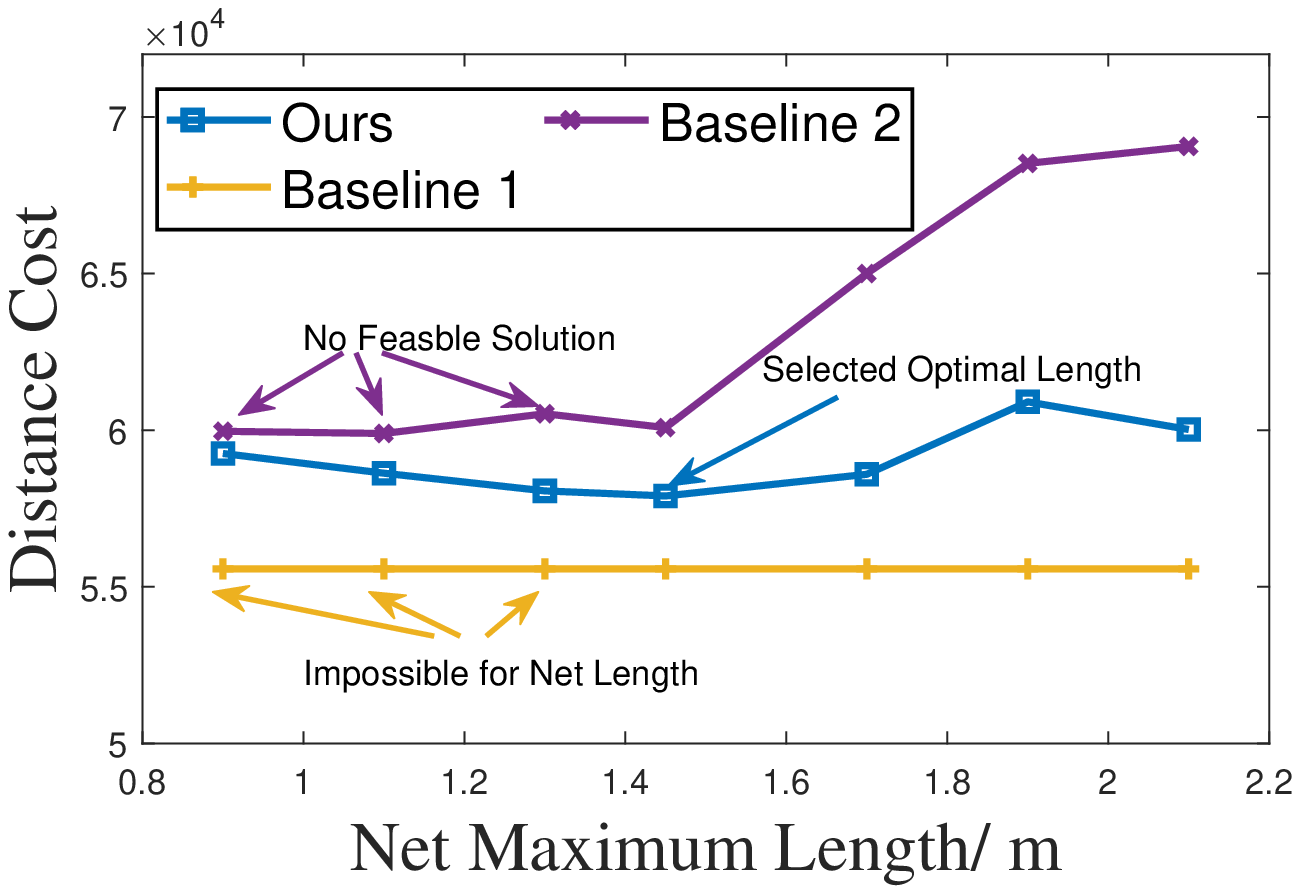}
        \caption{Shape cost comparison}
        \label{fig:compare_shape_cost}
    \end{subfigure}%
    \\%
    \begin{subfigure}[b]{0.5\linewidth}
        \centering
        \includegraphics[width=\linewidth,trim=4.4cm 10.5cm 4cm 10cm,clip]{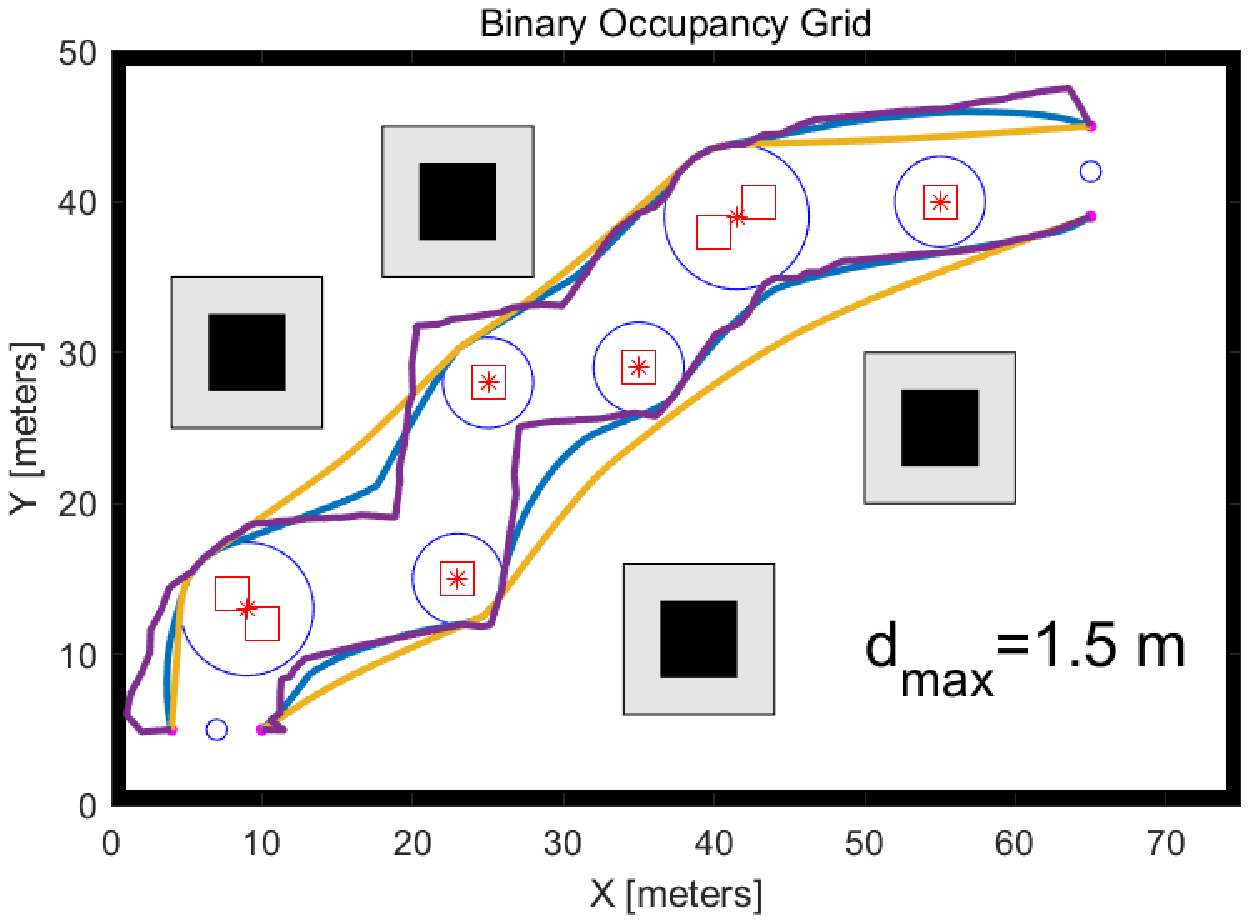}
        \caption{Trajectory comparison} 
        \label{fig:compare_shape_traj2}
    \end{subfigure}%
    \begin{subfigure}[b]{0.5\linewidth}
        \centering
        \includegraphics[width=\linewidth,trim=4cm 10.3cm 3.8cm 10.4cm,clip]{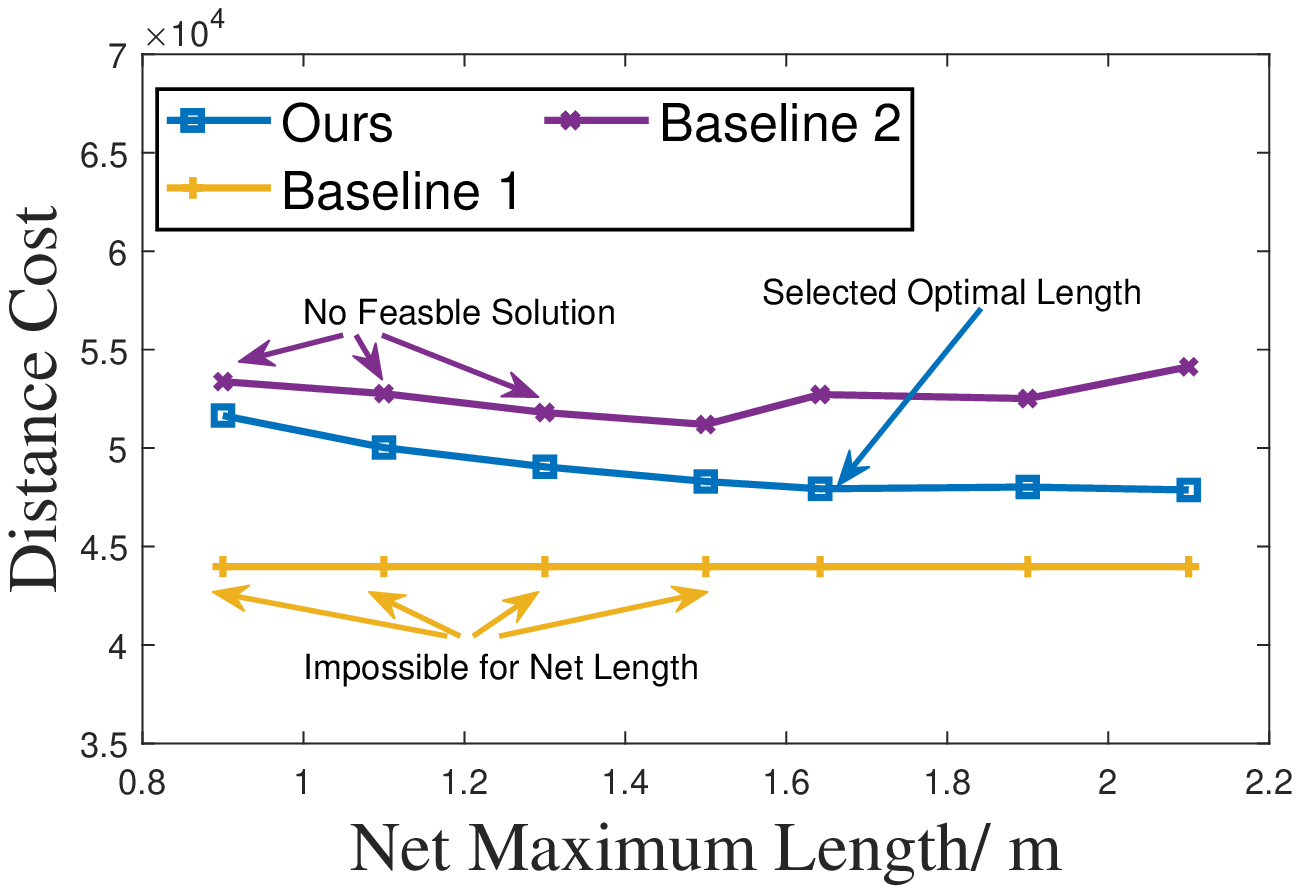}
        \caption{Shape cost comparison}
        \label{fig:compare_shape_cost2}
    \end{subfigure}%
    \\
    \begin{subfigure}[b]{0.5\linewidth}
        \centering
        \includegraphics[width=\linewidth,trim=4.4cm 10.5cm 4cm 10cm,clip]{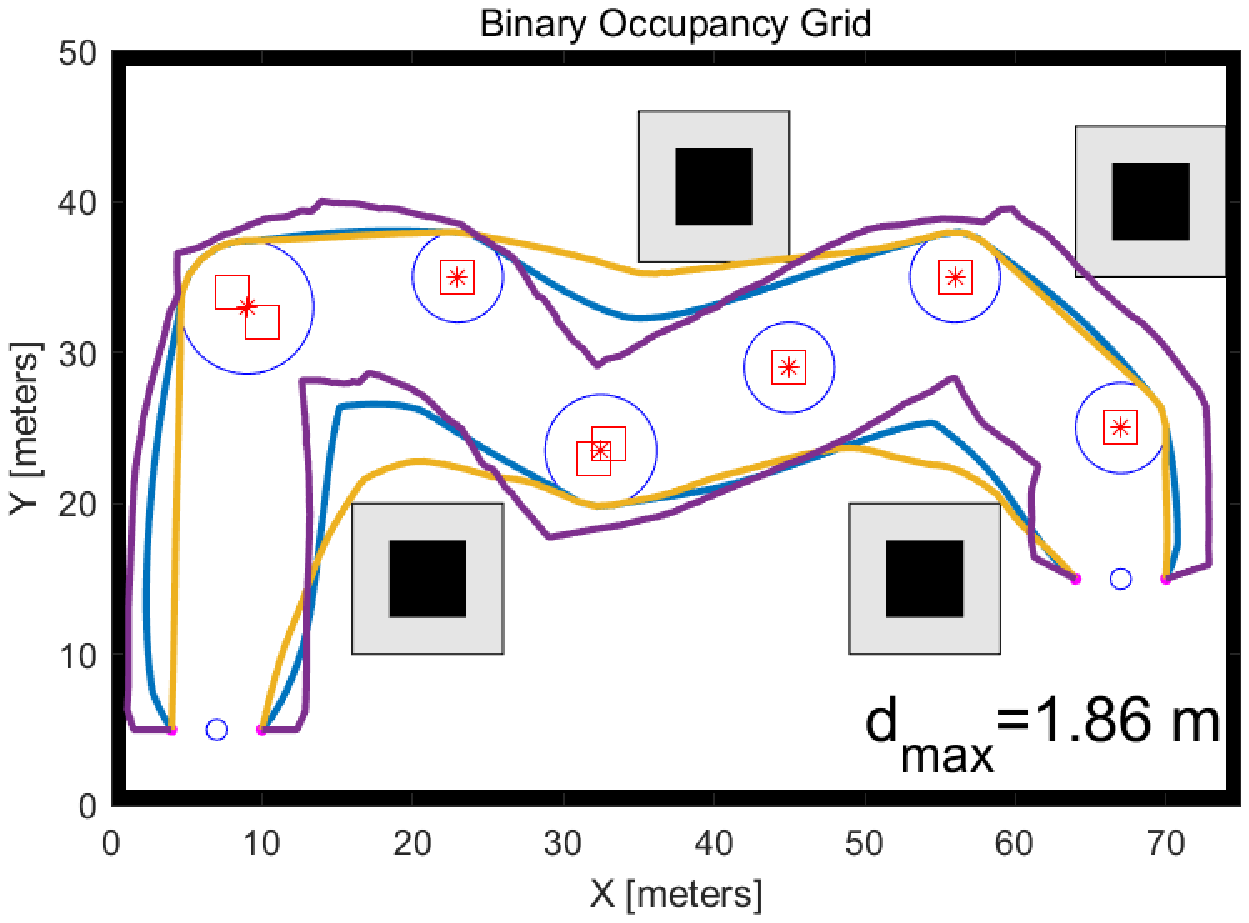}
        \caption{Trajectory comparison}  
        \label{fig:compare_shape_traj3}
    \end{subfigure}%
    \begin{subfigure}[b]{0.5\linewidth}
        \centering
        \includegraphics[width=\linewidth,trim=4.0cm 10.3cm 4cm 10.4cm,clip]{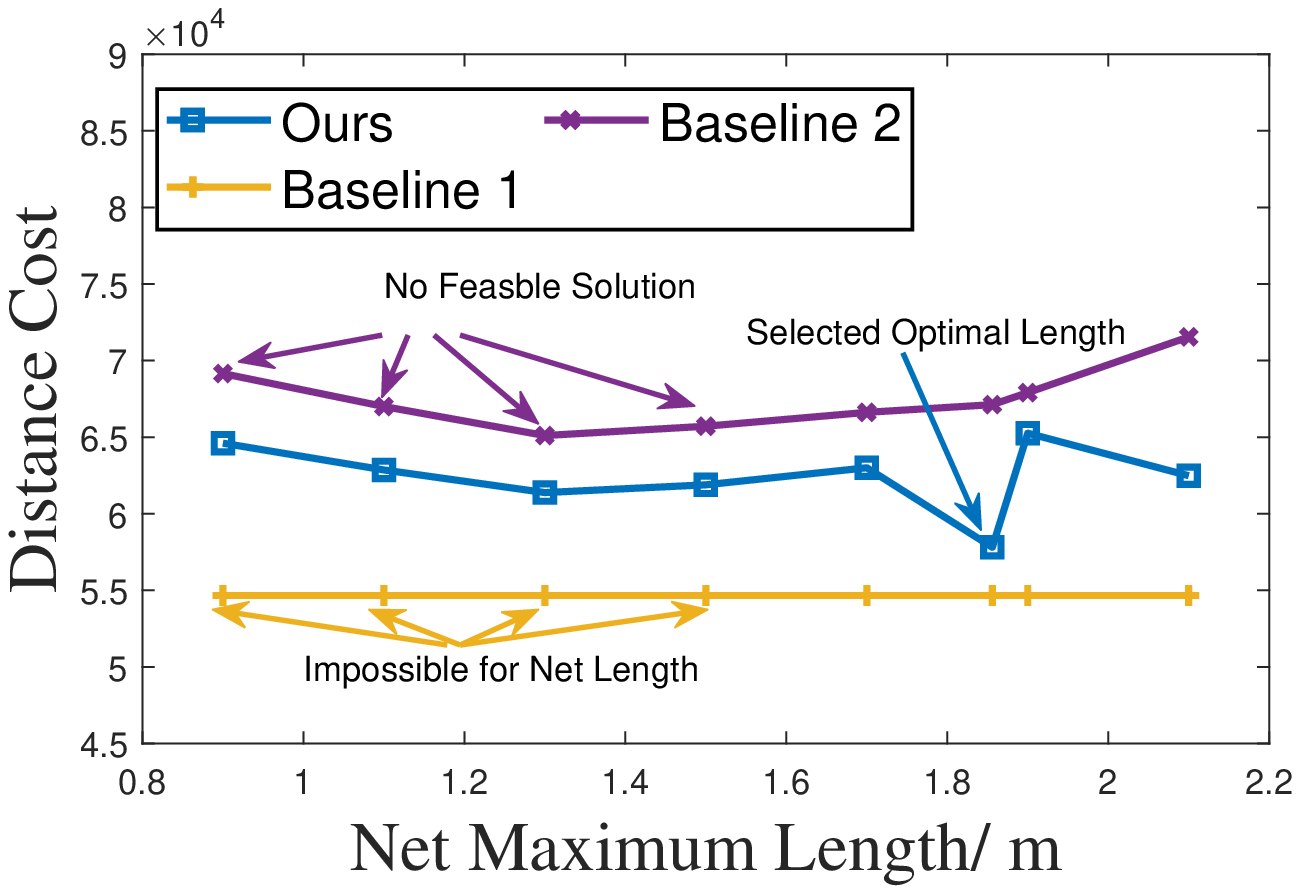}
        \caption{Shape cost comparison}
        \label{fig:compare_shape_cost3}
    \end{subfigure}%
    \caption{Planned trajectories with various expansion cost function designs in three scenarios; each row corresponds to one scenario.}
    \label{fig:cost_function}
\end{figure}

We design a U-shape cost function in \cref{eq:expan_cost} to penalize the tethered robot duo for being too close/far, so that the curvature of the net is indirectly controlled to embrace new objects and carry gathered objects. To demonstrate the efficacy of the U-shape design, we compare with two baseline cost functions:

\begin{itemize}[leftmargin=*,noitemsep,nolistsep,topsep=0pt]
    \item \textbf{No separation constraint included}. This strategy does not take separation distance into consideration~\cite{bhattacharya2015topological}; referred to as Baseline 1 in \cref{fig:cost_function}.
    \item \textbf{Desired separation distance $d_{\text{rest}}$ as hard constraint}. This strategy maintains the desired separation distance along the path~\cite{d2021catenary,talke2018catenary,laranjeira2017catenary,laranjeira2020catenary}; referred to as Baseline 2 in \cref{fig:cost_function}.
\end{itemize}

\cref{fig:compare_shape_cost,fig:compare_shape_cost2,fig:compare_shape_cost3} quantitatively compare the distance cost of each cost design with various net lengths in three scenarios, and  \cref{fig:compare_shape_traj,fig:compare_shape_traj2,fig:compare_shape_traj3} show the corresponding reference trajectories produced by our framework using the 3 types of expansion cost designs with a specific net length in each scenario. Although Baseline 1 yields the shortest travel distance without including any constraint on separation distance, it may require unrealistic net length. Baseline 2 keeps a specified separation distance at all time, which may find no solution for a small $d_{\text{max}}$ or cause unnecessarily long travel distance for a large $d_{\text{max}}$. In comparison, the proposed U-shape cost function produces feasible solutions for various $d_{\text{max}}$ while properly maintaining efficient travel distances. 

\section{Experiment}\label{sec:experiment}

\cref{sec:sim} has evaluated the proposed cooperative planning framework and the \ac{mrac} controller design in simulation. This section further validates the proposed framework in the physical environment.

\subsection{Setup}

The setup of our proposed tethered robot duo is shown in \cref{fig:moti_setup}, wherein a net is connected to two TurtleBot3 Waffle Pi robots at each end. Each TurtleBot is equipped with a Raspberry Pi 3B+ running \ac{ros}, an OpenCR 1.0 board with an IMU module and a microprocessor for low-level motor control, an LDS-01 Lidar for localization, and two Dynamixel XM430 motors with a maximum torque of 3 $N\cdot m$ for wheel actuation. The path planning component runs on a desktop (AMD Ryzen9 5950X CPU, 64.00 GB RAM) and takes an average of $180.3~s$ to produce the optimal trajectory with $N=200$ in \cref{alg:traj_opt}. 

\begin{figure}[t!]
    \centering
    \begin{subfigure}[b]{\linewidth}
        \centering
        \includegraphics[width=\linewidth,trim=0cm 0cm 0cm 3cm,clip]{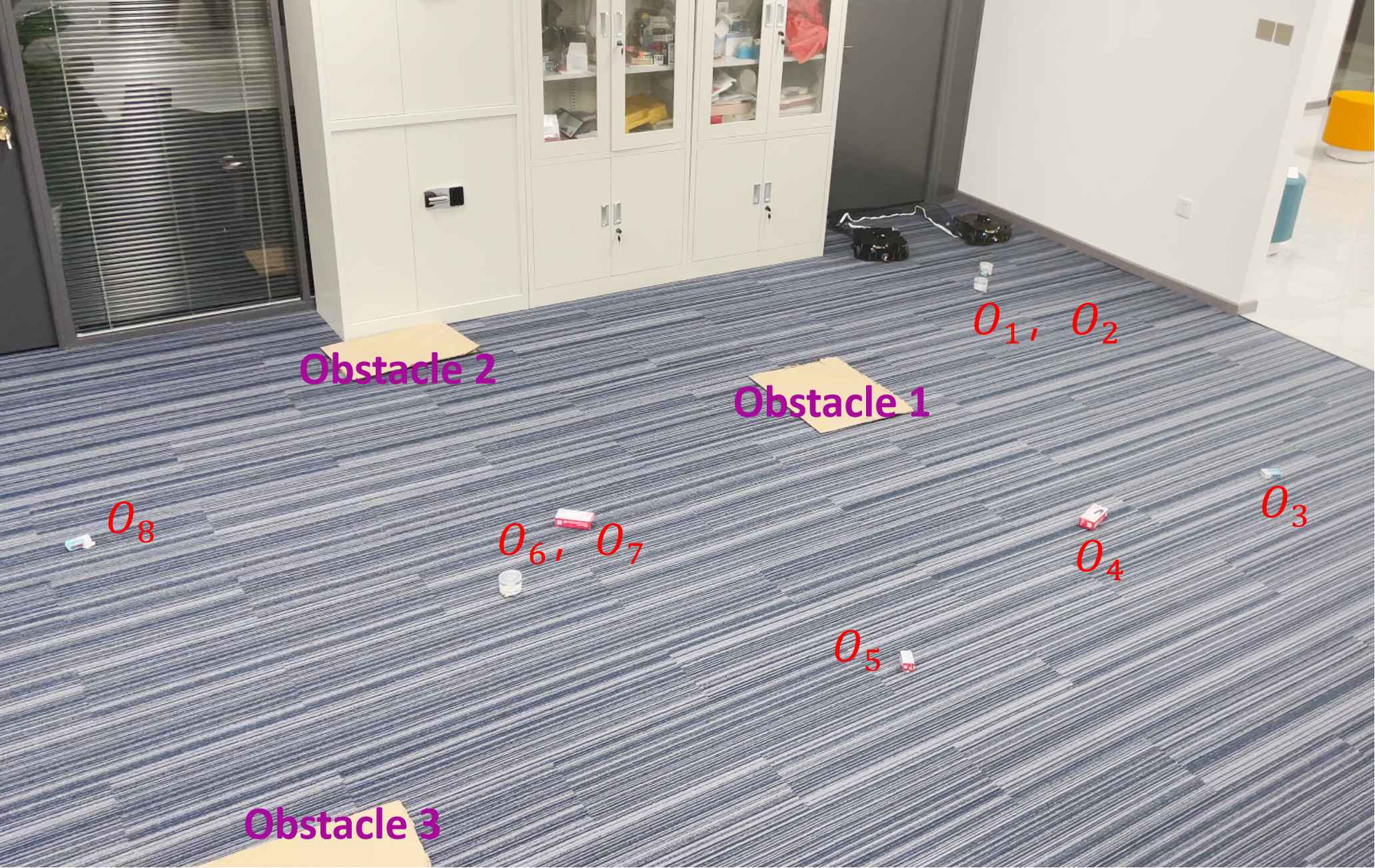}
        \caption{\textbf{Experimental environment.} Eight objects are placed on the floor. The robot duo is tasked to gather them with a tethered net while avoiding obstacles during its navigation.}
        \label{fig:setup}
    \end{subfigure}\\
    \begin{subfigure}[b]{\linewidth}
        \centering
        \includegraphics[width=\linewidth]{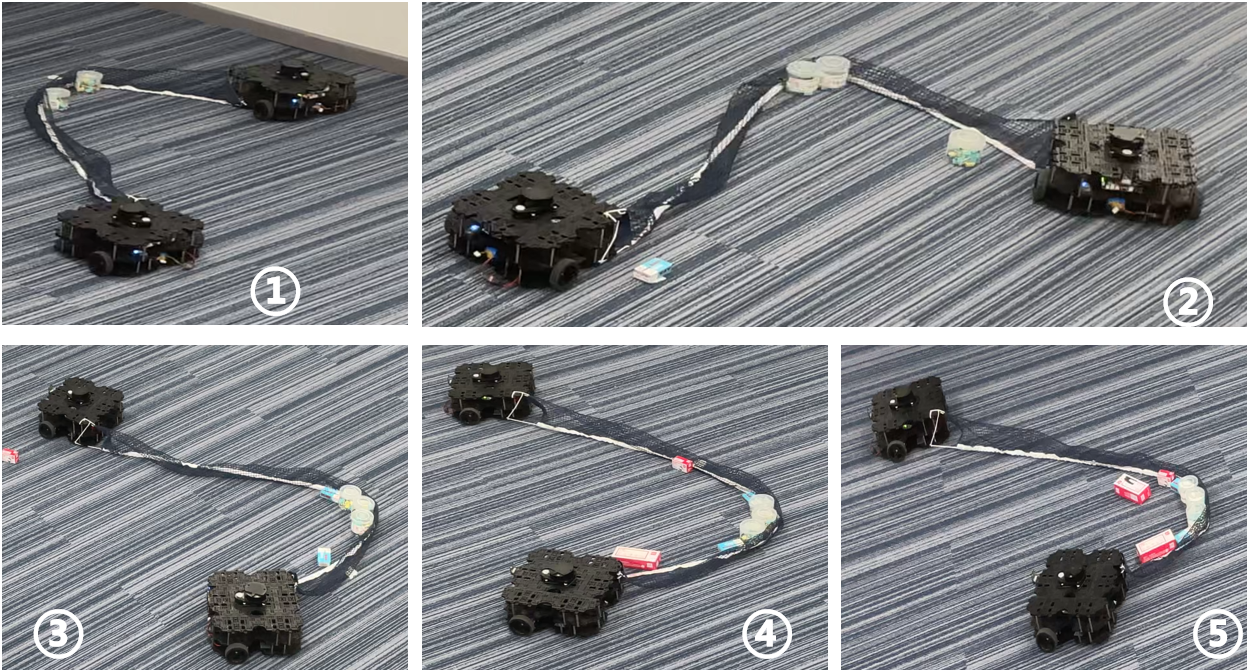}
        \caption{Five keyframes of gathering objects captured with varying weights during the robot duo's task execution.}
        \label{fig:exp_clips}
    \end{subfigure}
    \caption{Experiments with \ac{mrac} implementation in a robot duo in the physical environment.}
    \vspace{-6pt}
\end{figure}

To enable torque command, we modify the built-in low-level controller on OpenCR board designed to take velocity command inputs, such that it receives torque command inputs. The Gmapping \ac{slam} algorithm is utilized to localize the \acp{wmr} in experiment and outputs $q$ and $v$ as feedback. \cref{fig:setup} shows the experimental environment, a replication of the simulated environment presented in \cref{fig:compare_shape_traj}; eight objects are placed on the floor with various weights ranging from $15~g$ to $50~g$.

\begin{figure}[t!]
    \centering
    \includegraphics[width=0.85\linewidth,trim=4.9cm 11cm 4.5cm 10.8cm, clip]{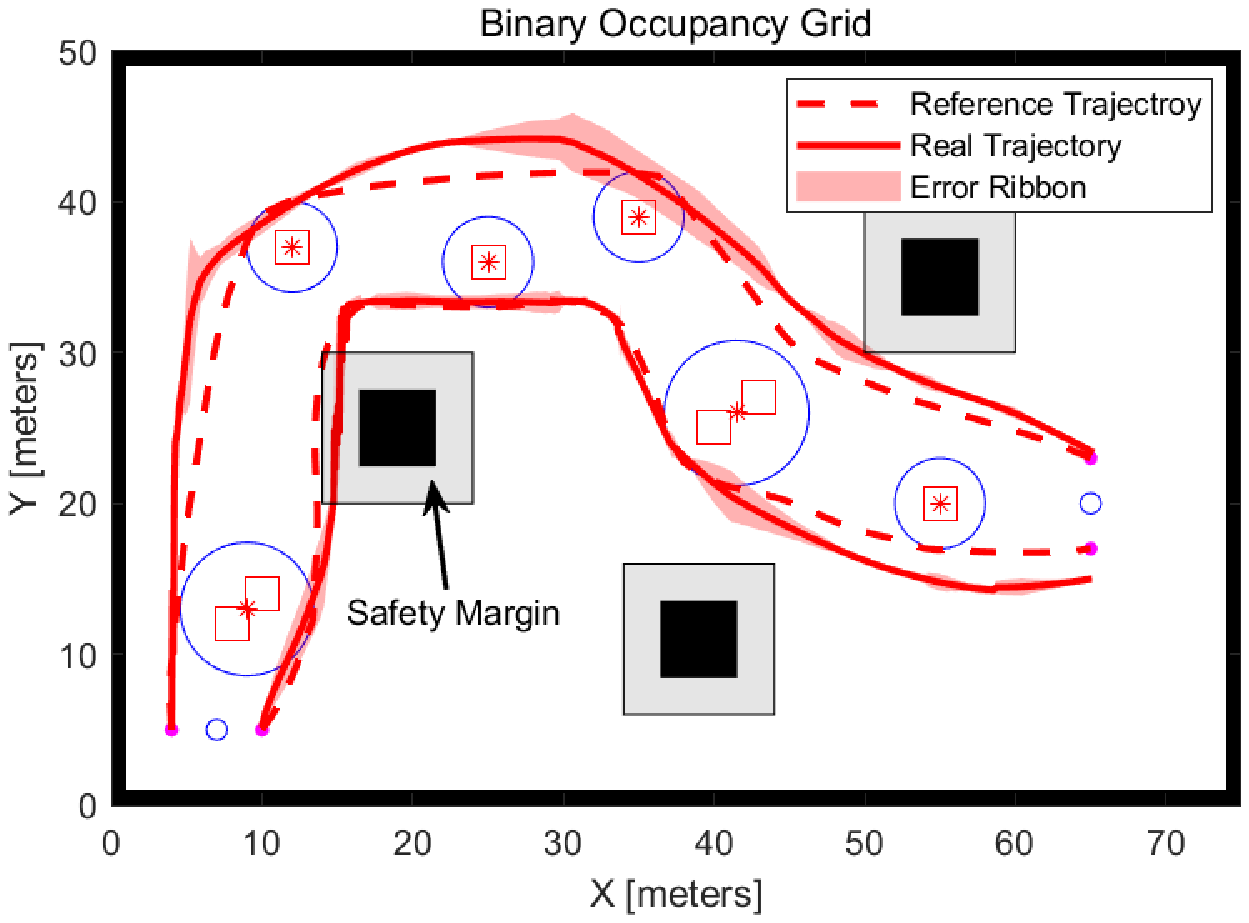}
    \caption{\textbf{Reference and tracked trajectories of the tethered robot duo.} The red shadows indicate the variations of five executions.}
    \vspace{-6pt}
    \label{fig:exp_tracking}
\end{figure}

\subsection{Results}

The net length used in our experiment is $1.45~m$, the optimal length estimated by our cooperative path planner according to this specific environment configuration. \cref{fig:exp_tracking} plots the reference and actual trajectories with error ribbon aggregated from five trials. Together with the simulation results, our experiments have demonstrated (i) the implemented \ac{mrac} controller robustly tracks the planned trajectories, and (ii) the tethered robot duo, using the proposed cooperative planning framework, successfully gathers all the objects along the path and carries them to the end position.

\section{Conclusion}\label{sec:conclusion}

This paper investigated an interesting task of gathering and transporting scattered objects with a tethered robot duo. We addressed it by proposing a two-stage cooperative planning framework based on trajectory optimization and implementing it with \ac{mrac}. In planning, we designed a U-shape cost function and incorporated other constraints to produce trajectories, capable of indirectly maintaining the flexible net's shape \wrt. the distance between two robots. In the implementation, we demonstrated that \ac{mrac} could robustly handle the increased payload with unknown dynamics as more objects were carried. As an extra feature, our planning framework can also estimate the most efficient length given an environment configuration, which led to a crucial extension to existing work in tethered robots that assumed an infinite or fixed tether length. Both the simulation and experiment results have demonstrated the efficacy of the proposed framework and the necessity of \ac{mrac} implementation. In future work, we plan to (i) extend this framework to tethered \ac{usv} and drones for collaborative tasks, (ii) incorporate a perception module when the environment configuration is not fully available, and (iii) enable effective re-planning for the robot duo to account for new observations.    

\section{Acknowledgement}
The authors would like to thank Dr. Bo Dai, Mr. Ziyuan Jiao, Mr. Aoyang Qin, and Ms. Zhen Chen at BIGAI for their constructive discussions and help on experiments and figures.

\bibliographystyle{ieeetr}
\bibliography{reference}

\end{document}

%% file: configs.tex
\usepackage{graphicx} \graphicspath{ {figures/} }
\usepackage{amsmath,amssymb,mathabx}
\usepackage{algpseudocode}
\usepackage[ruled]{algorithm2e}
\usepackage{acronym}
\usepackage{enumitem}
\usepackage{booktabs}
\usepackage{hyperref}
\usepackage[stable]{footmisc}
\usepackage{balance}
\usepackage{xspace,setspace}
\usepackage[skip=3pt,font=small]{subcaption}
\usepackage[skip=3pt,font=small]{caption}
\usepackage[dvipsnames]{xcolor}
\usepackage[capitalise]{cleveref}
\usepackage{tabularx,colortbl,multirow,array,makecell}
\usepackage{overpic}
\usepackage{cite}
\usepackage{tikz}

\makeatletter
\DeclareRobustCommand\onedot{\futurelet\@let@token\@onedot}
\def\@onedot{\ifx\@let@token.\else.\null\fi\xspace}
\def\eg{\emph{e.g}\onedot} 
\def\ie{\emph{i.e}\onedot} 
 
\def\etc{\emph{etc}\onedot} 
\def\wrt{w.r.t\onedot} 
\def\etal{\emph{et al}\onedot}
\makeatother

\crefname{algocf}{alg.}{algs.}
\Crefname{algocf}{Algorithm}{Algorithms}

\makeatletter
\def\BState{\State\hskip-\ALG@thistlm}
\makeatother

\makeatletter
\renewcommand{\paragraph}{%
  \@startsection{paragraph}{4}%
  {\z@}{0ex \@plus 0ex \@minus 0ex}{-1em}%
  {\hskip\parindent\normalfont\normalsize\bfseries}%
}
\makeatother

\crefname{algocf}{alg.}{algs.}
\Crefname{algocf}{Algorithm}{Algorithms}

\acrodef{dof}[DoF]{Degree of Freedom}
\acrodef{wmr}[WMR]{Wheeled Mobile Robot}
\acrodef{mrac}[MRAC]{Model Reference Adaptive Controller}
\acrodef{ros}[ROS]{Robot Operating System}
\acrodef{slam}[SLAM]{Simultaneous Localization and Mapping}
\acrodef{usv}[USV]{Unmanned Surface Vehicle}